\documentclass{article}

\usepackage{arxiv}

\usepackage[utf8]{inputenc} 
\usepackage[T1]{fontenc}    
\usepackage{hyperref}       
\usepackage{url}            
\usepackage{booktabs}       
\usepackage{amsfonts}       
\usepackage{nicefrac}       
\usepackage{microtype}      

\usepackage[english]{babel}
\usepackage{amsthm}
\usepackage{stackengine}[2013-10-15]

\usepackage{amsmath,amssymb}
\DeclareMathOperator*{\argmax}{argmax} 
\DeclareMathOperator*{\argmin}{argmin}
\usepackage[thinlines]{easytable}
\usepackage{csquotes}

\usepackage{natbib}
 \bibpunct[, ]{(}{)}{,}{a}{}{,}%

\usepackage{graphicx}

\usepackage{chngcntr}
\usepackage{apptools}

\usepackage{bbm}
\usepackage{enumerate}
\usepackage[shortlabels]{enumitem}

\title{Same-Day Delivery with Fair Customer Service}

\author{
    Xinwei Chen \\
    Department of Analytics \& Operations Management\\ Bucknell University\\
    Lewisburg, United States \\
    \texttt{xinwei.chen@bucknell.edu} \\
    \And
    Tong Wang \\
    Department of Business Analytics \\
    University of Iowa\\
    Iowa City, United States \\
    \texttt{tong-wang@uiowa.edu} \\
    \And
    Barrett W. Thomas \\
    Department of Business Analytics \\
    University of Iowa\\
    Iowa City, United States \\
    \texttt{barrett-thomas@uiowa.edu} \\
    \And
    Marlin W. Ulmer \\
    Otto-von-Guericke Universität Magdeburg\\
    Chair of Management Science\\
    Magdeburg, Germany \\
    \texttt{marlin.ulmer@ovgu.de} \\
    }

\begin{document}
\maketitle

\begin{abstract}
The demand for same-day delivery (SDD) has increased rapidly in the last few years and has particularly boomed during the COVID-19 pandemic. The fast growth is not without its challenge. In 2016, due to low concentrations of memberships and far distance from the depot, certain minority neighborhoods were excluded from receiving Amazon's SDD service, raising concerns about fairness. 
In this paper, we study the problem of offering fair SDD-service to customers. The service area is partitioned into different regions. Over the course of a day, customers request for SDD service, and the timing of requests and delivery locations are not known in advance. The dispatcher dynamically assigns vehicles to make deliveries to accepted customers before their delivery deadline. In addition to overall service rate (\textit{utility}), we maximize the minimal regional service rate across all regions (\textit{fairness}). We model the problem as a multi-objective Markov decision process and develop a deep Q-learning solution approach. We introduce a novel transformation of learning from rates to actual services, which creates a stable and efficient learning process. Computational results demonstrate the effectiveness of our approach in alleviating unfairness both spatially and temporally in different customer geographies. We also show this effectiveness is valid with different depot locations, providing businesses with opportunity to achieve better fairness from any location. Further, we consider the impact of ignoring fairness in service, and results show that our policies eventually outperform the utility-driven baseline when customers have a high expectation on service level. 
\end{abstract}


\section{Introduction}
Same-day delivery (SDD) is a service that enables customers to order and receive goods on the same day. It mimics the immediate product availability of a brick-and-mortar store with the convenience of ordering from electronic devices and home delivery \citep{hausmann2014same}. The SDD market has seen a fast growth over the past decade. These services have further gained popularity during the COVID-19 pandemic as they support social distancing while providing daily essentials to customers \citep{Perez,cnbc}. As a result, the growth has accelerated \citep{Perez2}. 

While the market is growing, the industry has been battling concerns about the fairness of these delivery services. In 2016, Amazon was accused of excluding certain minority neighborhoods from its SDD map \citep{howland}. Following criticism, Amazon cited two main reasons as to why those neighborhoods were excluded: low concentrations of Prime members and long distance to their warehouse \citep{ingold_2016}. Nonetheless, concerns about social justice, particularly given the vital role that delivery services such as Amazon Prime played during the pandemic, may arise if the population in the underserved neighborhoods consists of mostly certain race or age groups. Yet, no research has emerged to address this question.



Motivated by the difficulties cited by Amazon and the important role that SDD played for many people during the pandemic, we seek to fill the gap. Our first task is determining how to measure fairness and then how to optimize such that companies' need for financial return is balanced with fairness. To this end, we focus on geographical fairness in SDD service availability. That is, in this work, we partition the service area and measure fairness as the minimum service rate across the resulting regions. This measure of fairness reflects that it might be impossible and potentially undesirable to try to achieve equal service rates across all regions. Notably, equality is most easily achieved by serving fewer customers overall, and in fact, we can also achieve trivial equality by serving no customers at all, an outcome that serves neither companies nor their customers well. 

We study a SDD problem where customers from different designated regions make delivery requests throughout the course of the day. A central decision maker chooses which customers to serve and what vehicle should serve the customer request. With fairness in mind, we model the problem as a multi-objective Markov decision process of maximizing a weighted combination of expected overall service rate (\textit{utility}) and minimal regional service rate (\textit{fairness}). 



Both conceptually and methodologically, it is not straightforward to improve fairness and utility at the same time. The two measures are in competition. This challenge is further increased in a dynamic vehicle routing problem---decisions impact not only the current customer request and dispatch of vehicles, but the availability of resources in the future. Thus, it is particularly important for a solution approach to anticipate future demand so that myopic near-term decision making does not lead to poor outcomes later in the day. Additionally, SDD requires immediate response to customers, so a solution approach should also avoid time-consuming re-optimizations. For these reasons, we implement Q-learning to learn the values of state-action pairs. To overcome the large dimensionality of the state space, we use a neural network to approximate the Q-values. The use of the neural net allows the policy to be trained offline and thus does not require real-time re-optimizations.

At the same time, while both measures are naturally represented by rates, it is difficult to train the neural networks with the objective as a combination of rates. The challenge is that the denominator changes at each step. To avoid the problem of optimizing a rate, existing research focuses on maximizing the utility or the number of services per day rather than the service rate. However, because fairness requires a comparison across regions with potentially different demand rates, this transformation is less clear for fairness. We introduce a novel transformation that allows us to incrementally improve fairness using units common with the utility and thus create a stable and efficient learning process.

Using test sets designed to challenge one's ability to balance utility and fairness, we compare our proposed approach to a baseline that maximizes expected daily utility as well as two benchmark policies that use alternative mechanisms to those we propose to achieve fairness. Our results show that the proposed method is far superior to the benchmarks in terms of both utility and fairness while also achieving greater fairness with a minimal loss of daily utility relative to the baseline. However, we are also able to demonstrate that, in the long run, balancing utility and fairness can improve overall financial performance by maintaining customer loyalty.


Our work makes several contributions to the literature: 
\begin{itemize}
    \item \emph{Problem-oriented:} It is among the first to address social issues such as fairness in a dynamic vehicle routing problem. Our consideration of geographical fairness in services directly reflects the societal problems that neighborhoods that are financially advantageous or geographically closer to facilities receive more public or private resources, while the vulnerable with a disadvantage of location has a far lower chance of receiving the service.
    \item \emph{Method-oriented:} It is the first to solve a multi-objective dynamic vehicle routing problem using reinforcement learning. The incremental approach that we propose sequentially optimizes for fairness. Computational results demonstrate the effectiveness of the approach in alleviating the unfairness in SDD service opportunities for customers, with a small loss in utility compared to a highly utility-efficient benchmark from the literature. 
    \item \emph{Business-oriented:} We offer a tool for service providers to trade-off between utility and fairness. The effectiveness of our approach is valid with different depot locations, providing businesses with opportunity to achieve better fairness from any location. We also discover that ignoring fairness in geographical service in SDD has long-run business impacts as a result of alienating customers in underserved regions. This finding demonstrates that serving a broader set of customers than would otherwise have been served by an objective focused on daily utility offers the potential to improve a firm's long-run financial performance. 
\end{itemize}

The paper is organized as follows. In Section~\ref{sec:lit_review}, we present the literature related to SDD, fairness in vehicle routing problems, and fairness in machine learning. Section~\ref{sec:problem_model} describes the problem and presents a multi-objective Markov decision process model of the problem. In Section~\ref{sec:method}, we introduce the deep Q-learning solution approach for our problem. Section~\ref{sec:modified_objective} discusses the instability inherent in learning with service rates and introduces the modified objective. Section~\ref{sec:results} presents the details of experimental design and the results of our computational study. Section~\ref{sec:conclusions} closes the paper with conclusions and discussion on future work.



\section{Literature Review}\label{sec:lit_review}
In this section, we present the literature related to the SDDFCS. In Section~\ref{lit:sdd_fairness}, we review the literature of SDD. Section~\ref{lit:fairness_vrp} provides an overview of fairness related to vehicle routing problems (VRPs). In Section~\ref{lit:fairness_ml}, we present the fairness-related work in machine learning. 

\subsection{Same-Day Delivery}\label{lit:sdd_fairness}

As the market continues to grow, there is emerging literature on SDD. However, we are not aware of any that considers fairness in customer service. Instead, existing literature heavily studies utility-related objectives to improve the efficiency of the system \citep{azi,grippa,drones,liu_2019,klapp31,klapp32,klapp2020request,ulmer45,cosmi2019scheduling,cote2021dynamic,voccia,dayarian2020same,dayarian2020crowdshipping,ahamed2020deep,schubert2020same,Jahanshahi,bracher2021learning,deepQ}. 

Methodologically, the most related work to ours is \citet{deepQ}. In that work, the authors consider a SDD problem with a fleet of vehicles and drones and develop a deep Q-learning solution approach. However, they consider only a single objective that maximizes the expected number of customers served. In this paper, we consider a second objective of fairness in customer service, which as we discuss subsequently cannot be implemented directly and requires subsequent methodological development. Other work that uses deep reinforcement learning includes \citet{Jahanshahi} for a meal-delivery problem and \citet{ahamed2020deep} for crowdsourced urban delivery. Similar to \citet{deepQ}, their optimizations are driven by utility only. 

\subsection{Fairness in VRPs}\label{lit:fairness_vrp}

In this paper, we seek to incorporate fair customer service in a SDD system. Although fairness has not been studied in the SDD-literature, it has been explored in other fields such as financial services \citep{chen2012impact,zhu2012service}, telecommunication \citep{khan2012determinants,alzoubi2020perceived}, and in other VRPs. In this section, we focus on the fairness-related work in VRPs. It is worth mentioning here that the definition of fairness in machine learning is not the same as in the VRP-literature. In Section~\ref{lit:fairness_ml}, we will discuss fairness in machine learning. 




SDD falls into the category of dynamic and stochastic VRPs. In addition, the SDD literature is a subset of the dynamic routing literature. To the best of the authors' knowledge, customer fairness has also not been considered in the objective functions found in the dynamic routing literature. See \cite{ulmer2020modeling} for a review of the dynamic routing literature. 

In studying dynamic delivery problems, \citet{ninja_fairness} and \citet{marlin_fairness} consider fairness as an evaluation metric and highlight that the methods select lucrative customers in central areas and discriminate customers in more rural areas, especially in ``high-quality'' policies. In a restaurant-meal-delivery problem, \citet{ulmer2021restaurant} consider the fairness for drivers and use the average minimum and maximum numbers of services to evaluate the performance of policies. However, the authors in these papers do not optimize for fairness. Our work directly considers fairness for customers in the optimization. 


The concept of fairness has been explored in other VRPs, particularly in ride-sharing systems, where a fleet of vehicles picks up and drops off customers. Most problems in this field are modeled as an online bipartite matching problem \citep{ridesourcing,ridesharing_peakhour,ma2020group}. However, we are not aware of any work that directly optimizes for fairness as we do. \citet{chen_fairness} study a last-mile transportation system of vehicles that transport passengers from a train station to different destinations. The passengers are categorized by a regular or special type. They define two notions of fairness that consider the fare and the order of service for the two types of passengers. Instead of optimizing for fairness, the authors only constrain their solution with the two fairness notions. \citet{ridesourcing} investigate fairness in terms of driver profit. The authors measure the fairness as the minimum profit (utility) among all drivers, while our work in this paper considers the fairness for customers. In addition, \citet{ridesourcing} only improve the solution from an existing assignment until it satisfies a pre-defined threshold of fairness. \citet{ridesharing_peakhour} consider a ride-sharing problem during peak hours. There are different types of arriving requests, and the authors maximize the minimum match (service) rate over all types. Similar to ours, their measure also considers the group-level fairness for customers. However, similar to \citet{ridesourcing}, the authors develop a LP-based algorithm that improves the solution from two benchmarks. Further, the problem in this research is far more complex than bipartite matching, and thus we implement Q-learning to overcome the ``curses of dimensionality.'' 
In their work, \citet{ma2020group} also consider a measure of group-level fairness and improve their solution from benchmarks similar to \cite{ridesharing_peakhour}. 

While existing work that considers fairness in dynamic VRPs is scarce, fairness has been extensively studied for deterministic problems, especially within the public sector in operations research, such as school bus routing, disaster relief logistics, and healthcare deliveries. \citet{balcik} provide a thorough review of incorporating fairness in such domains with a perspective of vehicle routing. 
The measures of fairness in these papers include the maximum travel length of students \citep{bronshtein2014,pacheco2013,serna2001}, the minimum time loss of students due to taking the bus and waiting \citep{spada2005}, and the waiting- or arrival-time-related measures in disaster \citep{campbell2008,nolz2010,van2010,huang2012}. Work on workload fairness is also available in \citet{BowermanRobert1995,li2002school}, as well as in those from the literature on multi-objective VRPs \citep{jozefowiez2009evolutionary,wen2010dynamic,banos2013hybrid,kumar2014solving}. Balancing the workload can be seen as achieving fairness on the drivers' side, while our work considers fairness on the customers' side.

The problem examined in this paper is related but different from the VRPs discussed above. First, unlike the deterministic literature, we examine a dynamic and stochastic problem. Second, unlike the ride-sharing literature, we anticipate future requests rather than solve the problem in a rolling horizon. Finally, in contrast to the other fairness-related dynamic VRPs, we directly optimize for fairness, which requires a non-trivial change to objective function to facilitate learning within the deep Q-learning framework proposed in this paper. 

\subsection{Fairness in Machine Learning}\label{lit:fairness_ml}
Fairness in machine learning has received extensive interests in recent years. Research in this area has focused on combating the biases in the decision-making process, caused by bias in the training data or the algorithm. Interested readers are referred to \cite{mehrabi2019survey,barocas-hardt-narayanan} for reviews.

In machine learning, various definitions of fairness have been proposed in different contexts \citep{mehrabi2019survey}. There are mainly two types of bias that leads to unfair outcomes, data bias and algorithmic bias. Data bias arises when certain elements in the sampled data are over-representative \citep{olteanu2019social}. Algorithmic bias refers to the occurrence of unfair outcomes due to a direct consequence of the objective \citep{baeza2018bias}. Our work falls in the category of reducing algorithmic bias. 

To achieve fairness, existing research falls into three categories, pre-processing, in-process, and post-processing mechanisms. Pre-processing refers to learning a fair representation of the training data to remove the bias before training a model \citep{zemel2013learning, calmon2017optimized, madras2018learning,wangrepairing}. In-processing refers to directly learning a fair model through training \citep{berk2017convex}. Post-processing refers to making posthoc corrections to a biased pre-trained model \citep{hardt2016equality,lohia2019bias}. Our method in this paper falls into the second category, in-process mechanisms. To do that, we combine the multiple objectives of utility and fairness in a reinforcement learning model and directly optimize for a fair policy. 

In contrast to sequential decisions, most of the existing literature has focused on fair supervised learning applied to one-time predictions. However, there has been an increasing interest in fairness in a reinforcement setting. \cite{jabbari2017fairness} study fairness on actions (output) and find that, 
in theory, a learning algorithm requires an exponential amount of time in the number of states to achieve a non-trivial approximation of the optimal policy. In contrast, \cite{siddique2020learning} study fairness on users (input). 
They theoretically derive a new result in the standard reinforcement learning setting that provides a novel bound on the approximation error with respect to the optimal average reward and discounted reward.  
Our work in this paper focuses on a specific real-world problem. While, due to the dynamism and stochasticity in the problem, we are unable to measure performance relative to the optimal policy, we do some that, relative to benchmarks, our approach can achieve balance between utility and fairness.

\section{Problem Definition and Model}\label{sec:problem_model}
In this section, we formally present the problem of \textit{Same-Day Delivery with Fair Customer Service} (SDDFCS) and its model. Section~\ref{preliminary} first presents the problem narrative. In Section~\ref{preparation}, we discuss the preparation needed for the model. Then, in Section~\ref{mdp}, we model the SDDFCS as a multi-objective Markov decision process. The SDDFCS is similar to that found in \citet{deepQ}, but the objective is different and no drones are considered in this paper.

\subsection{Problem Narrative}\label{preliminary}
The SDDFCS is the problem of offering SDD in some service area. A dispatcher manages a fleet of vehicles that deliver goods to customers during the operating period. All vehicles start from the depot at the beginning of a day with no pre-assigned requests and must return to the depot by the end of the day. Packages can be loaded only at the depot. As a result, each vehicle is likely to have multiple delivery tours during the day and is assumed to deliver all the loaded packages before it can return to the depot. The dispatcher updates the routes for all vehicles in real-time throughout the day. 

Over the course of the day, customers make SDD requests. Their requesting time and location are not known until the request is made, but there is information available on the temporal and spatial distributions of requests. Upon receiving a request, the dispatcher needs to decide whether to offer SDD service, and if so, which vehicle will make the delivery for this customer. The decision is permanent in that, if offered the service, customers must be served by their delivery deadline. Customers that are not offered the service leave the system and are therefore ignored. 

\paragraph{Objective.} The objective in the existing literature is usually to maximize the \emph{overall service level} $r\textsubscript{total} \in [0,1]$, i.e., the expected number of accepted requests per day (called \emph{utility}) divided by the expected overall number of requests $n$. Previous work shows that, when maximizing service level, policies tend to favor customers who consume less resources \citep{deepQ}. These customers are typically those located close to other customers and/or close to the depot. In other words, the system is biased against customers who are located far from other customers or far from the depot, leading to unfair dispatching policies. 

In this paper, in addition to a high overall service level, we seek to achieve \textit{fairness} in service levels for customers from different geographic locations. To do so, we partition the service area into different regions, with some unfavorable than others. Our goal is not to keep the probabilities equal by all means because (as we show later) this could lead to poor service availability for everyone.

Instead, we measure the service level for all regions and define the fairness value as the \emph{minimal service availability across all regions}, $r_{\min} \in [0,1]$. We treat this value as an additional objective of the problem. Thus, our problem has two objectives, overall service level and minimum regional service level, weighted by a factor $\alpha\in[0,1]$:
\begin{equation}\label{eq:fake_objective}
    \max\ \mathbb{E}\big[(1-\alpha) \cdot r\textsubscript{total} + \alpha \cdot r_{\min}\big].
\end{equation}
The value of $\alpha$ indicates the different emphasis we put on fairness. When $\alpha=0$, no fairness is considered in the policy, so the policy is solely utility-driven as seen in most of the existing literature. When $\alpha=1$, the learning ignores utility and maximizes only the fairness measure. 

\subsection{Preparation of the Model}\label{preparation}

With the just described objective in mind, we model the SDDFCS as a multi-objective Markov decision process (MDP). In this section, we first introduce the general notion that is needed for the model and then define \textit{planned routes}, a concept that we use to model the updates and evolution of vehicles' routes. In Section~\ref{mdp}, we will formally present the multi-objective MDP.

The SDD service is offered in a service area $\mathcal{Z}$. The service area can be partitioned into $J$ regions, i.e., $\mathcal{Z} = \cup_{j=1}^{J} \mathcal{Z}_j$. On each day, during the time interval $[0, t\textsubscript{max}]$, a fleet of $M$ vehicles $\mathcal{V}=\{v_1,v_2,\dots,v_M\}$ loads packages from the depot $\mathcal{N}$ and delivers them to customers. The depot $\mathcal{N}$ is not necessarily located in the area $\mathcal{Z}$. Vehicles must return to the depot by $t_{\max}$. Due to the small size of most delivery items and the relatively small number of packages on most delivery routes, we assume vehicles are uncapacitated. Customers make delivery requests over the course of the day. We denote the set of customers by $\mathcal{C}$ and individual customers by $c_1$, $c_2$\dots. For a customer $c_k$, the time of request is $t(c_k)$. Accepted requests must be served within $\bar{\delta}$ units of time after the request is made. Thus, the delivery deadline of customer $c_k$ is $\delta(c_k)=t(c_k)+\bar{\delta}$. The travel times between two points are determined by function $\tau(\cdot , \cdot)$. It takes a vehicle $t_{\mathcal{N}}$ units of time to load packages at the depot and $t_{\mathcal{C}}$ units of time to drop off a package at a customer.

In the SDDFCS, vehicles perform repeated trips from the depot to a set of customers. Vehicles are not allowed to return to the depot unless they deliver all the loaded packages. To this end, we have two routes for every vehicle, an ongoing route that cannot be altered and a second ``planned'' route the vehicle will start after returning to the depot. Formally, we define a route that has not been started a \textit{planned route} \citep{ulmer2020modeling}. A vehicle's planned route contains the information on planned time of departure from the depot, customers to be served in this tour, times of arrival at each of the customers, and planned time of return to the depot. At $t=0$, the dispatcher starts to receive delivery requests and assign the accepted requests to vehicles' routes. These routes are planned routes because the vehicles still idle at the depot. 

Once a vehicle starts to load packages for the assigned customers and starts delivery, the planned route becomes an \emph{ongoing route}. Since no pre-emptive depot returns are allowed, the dispatcher is not allowed to integrate any new customers into any ongoing routes. Once a vehicle's planned route becomes an ongoing route, the dispatcher then constructs a new planned route for this vehicle. The dispatcher can assign new customers to this new planned route even when the vehicle is en route. This new planned route will be executed once the vehicle has returned to the depot after finishing its new ongoing route. 
Because the ongoing route cannot be altered, in the model, we need only to consider the return time from the ongoing route and the new planned route. Let $\Theta=\{\theta(v_1),\dots,\theta(v_M)\}$ represent the set of planned routes of all $M$ vehicles. A planned route is a sequence of customers (and the depot) to be visited with their locations and information on the arrival times. 
Because waiting at the depot is possible, a planned route also contains the planned time of departure from the depot. For vehicle $v_m$, its planned route is represented by
$$\theta(v_m)=((N_1^\theta,a(N_1^\theta),  s(N_1^\theta)),(c_1^\theta,a(c_1^\theta)),\dots,(c_h^\theta,a(c_h^\theta)),(N_2^\theta,a(N_2^\theta), t_{\max})).$$ 
The first entry of $\theta(v_m)$ represents the vehicle $v_m$'s return to the depot $N_1^\theta$ from the ongoing route. At $t=a(N_1^\theta)$, the vehicle returns to the depot. At $t=s(N_1^\theta)$, the vehicle $v_m$ starts to load packages for the customers assigned in this planned route, and this planned route becomes an ongoing route immediately. In the case it is the vehicle $v_m$'s first planned route in the day, the first entry represents the vehicle's initial position at the depot, and $a(N_1^\theta)=0$. The difference between $a(N_1^\theta)$ and $s(N_1^\theta)$ reflects the time the vehicle spends waiting at the depot before it starts its next delivery tour. Following the first depot visit is a sequence of customers $c_k^\theta$, $k=1,2,\dots,h$, that are assigned to vehicle $v_m$'s planned route but not yet loaded. Customer $c_k^\theta$ is planned to be visited at $t=a(c_k^\theta)$. The last entry in a planned route represents the vehicle's return to the depot at $t=a(N_2^\theta)$, implying the vehicle will wait until $t_{\max}$ if no additional customers are assigned later.  
\subsection{Multi-Objective Markov Decision Process Model}\label{mdp}

In this section, we present the multi-objective Markov decision process for the SDDFCS. Our model has two objectives, utility and fairness. The first objective is measured via $r\textsubscript{total}$ the overall service level of all customers, the latter is measured with respect to the region with minimal service level $r_{\min}$. 


\paragraph{Decision point.} A decision point is a time at which a decision needs to be made. In the SDDFCS, a decision point occurs when a customer makes a delivery request. Decision points are random, and we denote the $k^{th}$ decision point as $t_k$. It occurs when the $k^{th}$ customer $c_k$ makes a request at the time $t(c_k)$, and thus, $t_k=t(c_k)$.

\paragraph{State.} A state summarizes the information needed to make a decision when receiving the request from a customer $c_k$. Mathematically, we denote the state $S_k$ at decision point $t_k$ as a tuple $S_k=(t_k,c_k,\mathcal{C}_k^\Theta,\Theta_k,\Psi_k)$. The components are introduced in the following: 
\begin{enumerate}
    \item $t_k$: time of the decision point, $t_k=t(c_k)$.
    \item $c_k$: new customer request.
    \item $\mathcal{C}_k^\Theta$: set of customer requests that are accepted but not yet loaded at $t_k$. Note, these are also the customers that are present in the vehicles' planned routes at $t_k$. The delivery deadline for each customer in the set is also available.
    \item $\Theta_k$: set of planned routes for all the vehicles at $t_k$.
    \item $\Psi_k$: set of the total numbers of requests received and requests accepted preceding $t_k$ for all the regions, denoted by $\Psi_k=\{(\psi^{\textrm{total}}_{1,k},\psi^{\textrm{accept}}_{1,k}),\dots,(\psi^{\textrm{total}}_{j,k},\psi^{\textrm{accept}}_{j,k})\}$. Each value in the set can be updated by counting the number of requests and recording the value of $a_k$ (to be introduced next). Similarly, the total service rate can also be calculated using these numbers. 
\end{enumerate}

In the initial state $S_0$, all vehicles are available at the depot, and no customers are pre-assigned to them.

\paragraph{Action.} At every decision point $t_k$, we select an action $x_k$ from $\mathcal{X}_k$, the set of available actions in the state $S_k$. The action $x_k=(a_k,\Theta_k^x)$ consists of two parts, acceptance decision and routing decision. For a customer $c_k$, we first decide whether to accept the request, offering service to the customer. The acceptance decision $a_k$ is defined as:
$$ a_k=\left\{
\begin{array}{rcl}
0       &      & {\textrm{if $c_k$ is not offered service,}}\\
1     &      & {\textrm{if $c_k$ is offered service.}}
\end{array} \right. $$
Then, we make the routing decision. We define $\Theta_k^{x}$ as updated set of planned routes, and $\mathcal{C}_k^{\Theta,x}$ as updated set of accepted customers requests. If customer $c_k$ is not offered service, then $\mathcal{C}_k^{\Theta,x}=\mathcal{C}_k^{\Theta}$. If $c_k$ is offered service, we update the corresponding customer and planned-route sets to incorporate the addition of $c_k$. In such case, an update is feasible if it satisfies the following conditions:

\begin{enumerate}
     \item The planned routes in $\Theta_k^{x}$ contain $c_k$ and all the customers in $\mathcal{C}_k^{\Theta}$.
     \item For every customer in $\mathcal{C}_k^{\Theta,x}$, the planned arrival time is not later than the delivery deadline.
     \item For each vehicle $v_m$'s planned route in $\Theta_k^{x}$, its start time of loading is not earlier than the time at which it returns to the depot, $a(N_1)\leq s(N_1)$.
     \item In each planned route in $\Theta_k^{x}$, the difference between the start time of loading and the arrival time at the first customer is the sum of travel time and loading time.
     \item In each planned route in $\Theta_k^{x}$, the difference between the arrival times of two consecutive customers is equal to the sum of travel time and service time.
     \item Vehicles must return to the depot not later than the end of the day, $t\textsubscript{max}$.
\end{enumerate}
In Section~\ref{heuristic}, we will present the routing heuristic that we use to reduce the action space $\mathcal{X}$ to $\hat{\mathcal{X}}$.

\paragraph{Reward.} At decision point $t_k$, the change in the total service rate resulting from taking the action $x_k$ can be expressed as 
\begin{equation}\label{eq:fake_r_total}
    R\textsubscript{total,k}=\frac{ a_k +\sum_{j=1}^{J}  \psi^{\textrm{accept}}_{j,k}}{k}-\frac{\sum_{j=1}^{J} \psi^{\textrm{accept}}_{j,k}}{k-1},\ k\geq 2.
\end{equation}
The summation calculates the total number of requests accepted before $t_k$, and the value of $a_k$ results from the acceptance decision to the current request $c_k$. Similarly, the change in the minimal regional service rate is defined as 
\begin{equation}\label{eq:fake_r_min}
    R\textsubscript{min,k}=
    \min_{j\in 1,\dots,J}
    \frac{\mathbbm{1}_{\mathcal{Z}_j}(c_k)\cdot a_k+\psi^{\textrm{accept}}_{j,k}}
    {\mathbbm{1}_{\mathcal{Z}_j}(c_k)\cdot 1+\psi^{\textrm{total}}_{j,k}}
    -
    \min_{j\in 1,\dots,J}
    \frac{\psi^{\textrm{accept}}_{j,k}}
    {\psi^{\textrm{total}}_{j,k}},
    \ k\geq 2.
\end{equation}
The indicator function $\mathbbm{1}_{\mathcal{Z}_j}(c_k)$ is $1$ if request $c_k$ is made from region $j$, and $0$ otherwise. 

At decision point $t_1$, there are no previous requests yet. To avoid a zero denominator, we simply define $R\textsubscript{total,1}=a_1$, i.e., $1$ if request $c_1$ is accepted and $0$ otherwise. Similarly, we define $R\textsubscript{min,1}=0$ as the minimum regional service rate remains at $0$ regardless of which region request $c_1$ is made from. Thus, the (immediate) reward of an action $x_k$ given the state $S_k$ is a linear combination of the two changes, expressed as
\begin{equation}
    R(S_k,x_k)=(1-\alpha)\cdot R\textsubscript{total,k}+\alpha\cdot R\textsubscript{min,k}.
\end{equation}

\paragraph{Post-decision state.} After an action $x_k$ is made in the state $S_k$, a post-decision state $S^x_k$ occurs. The post-decision state $S^x_k$ contains the following information:

\begin{enumerate}
    \item $t_k$: Point of time.
     \item $\mathcal{C}^{\Theta,x}_k$: Set of customers still to be loaded and served as well as their deadlines. The update of the set $\mathcal{C}_{k}^{\Theta,x}$ is dependent on the acceptance decision $a_k$.
    \item $\Theta^x_k$: Updated set of planned routes for all vehicles.
    \item $\Psi^x_k$: Updated set of the total numbers of requests received and requests accepted for all the regions, reflecting the acceptance decision $a_k$. 
\end{enumerate}

\paragraph{Stochastic information and transition.} After the action $x_k$ is selected, the MDP proceeds until either a new customer $c_{k+1}$ makes a request or the MDP terminates. This is determined by the exogenous information of the process, and we denote the realized information after $x_k$ as $\omega_{k+1}$. If the realization is empty $\omega_{k+1}=\{\}$, the MDP terminates in the final state, i.e., $S_{k+1}=S_K$. If the realized exogenous information contains a customer request $\omega_{k+1}=\{c_{k+1},t(c_{k+1})\}$, the MDP transits to the next decision point $t_{k+1}=t(c_{k+1})$. The new state $S_{k+1}$ combines the information from the post-decision state and the realized exogenous information, which reflects the following updates:

\begin{enumerate}
    \item The time of the decision point $t_{k+1}$ is $t(c_{k+1})$, i.e., the time of the new request $c_{k+1}$.
    \item In the case of $t_{k+1}<s(N_1^{\theta})< t_{k+1}$ for some vehicle, the transition from $t_k$ to $t_{k+1}$ leaded to the transformation of its planned route to an ongoing one. As a result, a set of customers were loaded between $t_{k}$ and $t_{k+1}$. We denote the set of such customers by $\mathcal{C}^{\Theta,x}_{k,k+1}$. These customers are no longer part of the state, and therefore, $\mathcal{C}^{\Theta}_{k+1}=\mathcal{C}^{\Theta,x}_k \backslash \mathcal{C}^{\Theta,x}_{k,k+1}$. 
    \item Similarly, the set of planned routes $\Theta_{k+1}$ is updated by removing all delivery tours that already started between $t_{k}$ and $t_{k+1}$.
    \item Since no new requests are made until $t_{k+1}=t(c_{k+1})$, the acceptance information $\Psi_{k+1}$ is the same as $\Psi^x_k$.
\end{enumerate}

\paragraph{Objective.}

A solution to the SDDFCS is a policy $\pi\in\Pi$ that assigns an action to each state. The optimal solution $\pi^*$ is 
\begin{equation}\label{obj}
\pi^*=\argmax_{\pi \in \Pi}\ \mathbb{E} {\bigg[\sum_{k=0}^{K} R(S_k,X^\pi(S_k))|S_0\bigg]}.
\end{equation}
Note that $X^\pi(S_k)$ here represents the action selected given the state $S_k$ under the policy $\pi$. 

\section{Solution Approach}\label{sec:method}
To solve MDPs, one can perform backward induction to the Bellman equation 
\begin{equation}\label{eq:bellman}
V(S_k)=\max_{x\in \mathcal{X}_k}\{R(S_k,x)+\mathbb{E}[V(S_{k+1})|S_k]\}.
\end{equation}
However, due to the ``curses of dimensionality'' in the SDDFCS, backward induction is not computationally tractable. Specifically, there are several challenges our solution approach should address. First, the action space is large because an action needs to indicate whether the request is accepted, and if so, which vehicle to assign it to. The large action space makes it challenging to make a decision especially in real time. Second, the problem has a large state space as a state stores the information about the time of current decision point, the new customer request, the fleet of vehicles, and temporal overall and regional service levels. The challenge arises as how to efficiently process and evaluate the large amount of information extracted from the state. Finally, the additional consideration of fairness requires the objective to operate on service rates. However, as we will show in Section~\ref{sec:modified_objective}, it is not stable to directly optimize them. To overcome these challenges, we first implement a routing heuristic to reduce the action space. Then, we develop a deep Q-learning solution approach for the SDDFCS. For the instability in learning directly with rates, in Section~\ref{sec:modified_objective}, we modify the reward function and thus the objective.





\subsection{Routing Heuristic and Reduced Action Space}\label{heuristic}
The multi-objective MDP has a very large action space $\mathcal{X}_k$ in every state, particularly resulting from the update of planned routes introduced in Section~\ref{mdp}. It is well known that, even routing with only one vehicle, the number of feasible plans increases exponentially as more requests are assigned. Even off-line training such as Q-learning cannot cope with such a large action space. Further, the SDDFCS requires fast response to customers, so real-time optimizations are usually not practical especially with a fleet of multiple vehicles like ours. For these reasons, we decide about the assignment but reduce the action space using a routing heuristic to assign new customers to vehicles' planned routes. The insertion heuristic we use is similar to those from the literature \citep{azi,deepQ}. It seeks to insert a new customer into an existing route such that the insertion minimizes the increase in the tour time resulting from serving the additional customer, while still satisfying every customer's delivery deadline and the requirement of $t\textsubscript{max}$. 
     
With the heuristic, we reduce the original action space to $\hat{\mathcal{X}}$. The reduced action space $\hat{\mathcal{X}}$ consists of $M+1$ actions, i.e., $\hat{\mathcal{X}} = \{0, 1, \cdots, M\}$. For every $m \in \hat{\mathcal{X}}$ and $m>0$, action $m$ represents assigning the request to vehicle $v_m$. Action $0$ represents not offering service. At a decision point, the dispatcher iterates through all vehicles to determine the feasibility of serving the customer by each vehicle according to the six conditions described in Section~\ref{mdp}. We denote the feasibility of vehicle $v_m$ serving customer $c_k$ by an indicator variable $a_{m,k}$. The value of $a_{m,k}$ is $1$ if vehicle $v_m$ can feasibly serve $c_k$, and $0$ otherwise. Given the feasibility of vehicles, a set of feasible actions $\hat{\mathcal{X}_k} \subseteq \hat{\mathcal{X}}$ at the decision point $t_k$ is defined as: 
\begin{equation}
    \hat{\mathcal{X}_k}=\left\{
    \begin{array}{ccl}
        \{0\} &\ & {\textrm{if $a_{m,k}=0$\ \ $\forall m  =1,2,\dots, M,$}}\\
             \{0\}\cup\{m:a_{m,k}=1\}     &\ & {\textrm{otherwise.}}
    \end{array} \right. 
\end{equation}

\subsection{Deep Q-Learning for SDDFCS}\label{sec:q-value}
In the SDDFCS, current decisions impact the availability of delivery resources in the future and thus the overall performance of a policy in both utility and fairness. Q-learning tackles MDPs by learning the expected value of state-action pairs and thus satisfies the need to anticipate for the future in our problem. 

For a state $S_{k}$ and the action taken $x_{k}$, the Q-value function can be expressed as
\begin{equation}
    Q(S_k, x_k)=R(S_k,x_k) + \mathbb{E}[\hat{V}(S_{k+1})|x_k].
\end{equation}
The $\hat{V}(S_{k+1})$ is the value of the state $S_{k+1}$ when operating on the reduced action space $\hat{\mathcal{X}}$. With $Q(S_k, x_k)$ and the set of feasible actions $\hat{\mathcal{X}_k}$, we can solve an approximation of Bellman equation (Equation~\ref{eq:bellman}) expressed as
\begin{equation}\label{eq:qBellman}
\hat{V}(S_k)=\max_{x\in \hat{\mathcal{X}_k}}\{Q(S_k, x)\}.
\end{equation}

In addition to the action space, the SDDFCS has a large state space. To efficiently compute with the large amount of information, we build a neural network to approximate the Q-values for the SDDFCS. The neural network is defined by a set of weights $\phi$. 
The input layer takes the state as features. These features help us group states with similar values and differentiate states with different values. We select the features that summarize the information about the following:
\begin{itemize}
    \item \textit{Decision point}: the time of decision point. Research has shown the value of point in time in studying SDD problems \citep{ulmer2017approximate} and the monotonicity of the value function in time \citep{deepQ}.
    
    \item \textit{Customer request}: region where the requesting customer resides, and the vehicle travel time directly from the depot to the customer. These features help us locate the customer. 
    
    \item \textit{Fleet}: the time at which each vehicle returns to the depot from its ongoing tour, feasibility of each vehicle to serve the new customer, and if so, the corresponding increase in travel time. In addition to point in time, the value function is monotonically decreasing in the times at which vehicles return to the depot \citep{deepQ}. The increase in each vehicle's travel time reflects the impact of action selection on the fleet.
    
    \item \textit{Fairness}: temporal acceptance rates at the time of decision point for all regions. These features offer an indication of the impact of the decision-making history on the objective value at the current decision point.
\end{itemize}
We normalize all the features using \emph{min-max normalization}.

\section{Modified Objective}\label{sec:modified_objective}

When the objective is to optimize service rates, the MDP model needs to compute these rates at every decision point. However, due to the changing numbers of requests received and accepted in different states, it is challenging to operate on these rates. In this section, we first illustrate these challenges, and then introduce a modified objective that is more amenable to use with the proposed solution approach. 

\subsection{Instability of Learning Directly with Rates}\label{sec:instability}

In this section, we demonstrate the instability inherent in learning when the objective is composed of rates. As the goal is to improve both overall and minimal regional service rates, it seems straightforward to directly maximize the combination of these two rates as instructed in Section~\ref{mdp}, i.e., $(1-\alpha)\cdot r\textsubscript{total} + \alpha\cdot r\textsubscript{min}$. However, due to the changing denominator over the course of the day, it is difficult to do so. 
Specifically, the denominators in Equations~\ref{eq:fake_r_total} and~\ref{eq:fake_r_min} change at each state, resulting in an unstable learning mainly from two aspects, \emph{volatile combined reward} and \emph{inconsistent improvement in fairness}. We will illustrate both in the following.

\paragraph{Volatile combined reward. }Assume there are two regions, and the value of $\alpha$ is $0.5$. Suppose the dispatcher has accepted one out of one request from Region~$1$, and four out of five requests from Region~$2$, resulting in the current $r\textsubscript{total}=\frac{5}{6}$ and $r\textsubscript{min}=\frac{4}{5}$. The decision point $t_7$ occurs when customer $c_7$ makes a request from Region~$1$. If $c_7$ is offered the service, the reward of the action is $(1-0.5)\cdot (\frac{6}{7}-\frac{5}{6})+0.5\cdot (\frac{4}{5}-\frac{4}{5})\approx 0.012$. However, if not offering the service, the reward becomes $(1-0.5)\cdot (\frac{5}{7}-\frac{5}{6})+0.5\cdot (\frac{1}{2}-\frac{4}{5})\approx -0.210$. In general, the reward value will vary substantially in the beginning of the day and will be very volatile due to the small denominators.

\paragraph{Inconsistent improvement in fairness. }As the denominator becomes large enough later in the day, the impact of an action is almost negligible. For example, when three out of five requests are already accepted from the minimal-service-rate region, the acceptance to the sixth request improves $r\textsubscript{min}$ by $\frac{4}{6}-\frac{3}{5}=0.067$, while this improvement becomes only $0.003$ if $70$ out of $100$ requests are already accepted. This diminishing behavior is similar to what we see in previous example.

\begin{figure}[h!]
    \centering
	\begin{minipage}[b]{0.5\linewidth}
		\centering
		\includegraphics[width=\linewidth]{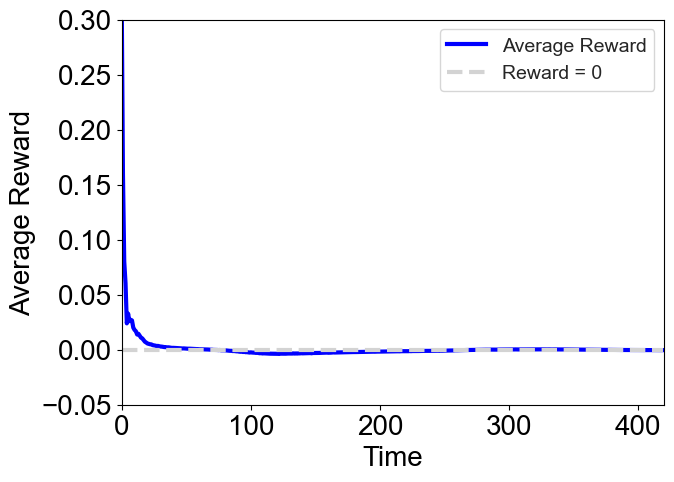}
		\end{minipage}
	\caption{Reward curve when learning with rates for one of computational settings and $\alpha=0.5$.}
	\label{fig:reward_learning_curves}
\end{figure}
Figure~\ref{fig:reward_learning_curves} presents the reward curve for a case when we apply a myopic policy. We calculate the average reward per minute over $500$ days. The horizontal axis represents time throughout the day, and the vertical axis is the reward calculated with $\alpha=0.5$ using Equations~\eqref{eq:fake_objective}, \eqref{eq:fake_r_total} and \eqref{eq:fake_r_min}. The computation confirms what the illustrative examples suggest---the reward is large in the beginning, becomes negligible thereafter, and eventually approaches zero. As a consequence, there is no feedback by the end as the marginal values have gone to zero. So whether accepting a customer or not has almost the same value. As we show later, these marginal rates impede the learning process substantially, especially given the approximation errors usually observed when learning.

\subsection{Modification of Objective}\label{sec:modified_reward}
To avoid the instability due to the changing denominators and the resulting lack of feedback later in the horizon, we propose a modified version of the reward function and thus the objective that draws on incremental improvement in utility and fairness.


To manage the expected \emph{utility},  maximizing the expected number of customers served is equivalent to maximizing the expected rate of customers served. Thus, we set the utility reward to $1$ if offering service, and $0$ otherwise. However, we must put the fairness in comparable units to the utility. How to do so is less obvious than with the utility. 

For the fairness part, to reduce the impact of changing denominator seen in Section~\ref{sec:instability}, we instead operate on a fixed denominator. For each region, we use $n_j$, the expected number of requests over the course of a day for region $j$. The aid of constant $n_j$ helps us avoid inconsistent improvements in $r\textsubscript{min}$. Accordingly, $n=\sum_{j=1}^{J} n_j$ denotes the expected total number of requests. Then, at each decision point $t_k$ and before action $x_k$ is made, we identify the index $z\textsubscript{min,k}$ for the region with the minimal regional service rate as $$z\textsubscript{min,k}=\argmin_{j=1,\dots,J}\frac{\psi^{\textrm{accept}}_{j,k}}{n_j}.$$ 
After action $x_k$ is made, the partial reward for fairness is then defined as 
$$R\textsubscript{min,k}=\frac{\mathbbm{1}_{\mathcal{Z}_{z_{\textrm{min,k}}}}(c_k)\cdot a_k}{n_{z_{\textrm{min,k}}}}.$$ 
The value of $R\textsubscript{min,k}$ is then $1/n_{z_{\textrm{min,k}}}$ if $c_k\in \mathcal{Z}_{z_{\textrm{min,k}}}$ and $a_k=1$, and $0$ otherwise. Thus, the (immediate) reward of an action $x_k$ given the state $S_k$ is 
defined as 

\begin{equation}\label{immediate_reward}
    R(S_k,x_k)=\left\{
    \begin{array}{lcl}
    1-\alpha+\alpha\cdot \frac{1}{n_{z_{\textrm{min,k}}}}\cdot n &\ & \textrm{if } a_k=1 \textrm{ and } c_k\in\mathcal{Z}_{z_{\textrm{min,k}}}, \\
    1-\alpha   &\ & \textrm{if } a_k=1\textrm{ and }c_k\not\in \mathcal{Z}_{z_{\textrm{min,k}}},\\
    0 &\ & \textrm{otherwise.}
    \end{array} \right . 
\end{equation}
With this formulation, reward is non-negative throughout the day, and both utility and fairness measures are consistently incremented over time. 

\begin{figure}[h!]
    \centering
	\begin{minipage}[b]{0.45\linewidth}
		\centering
		\includegraphics[width=\linewidth]{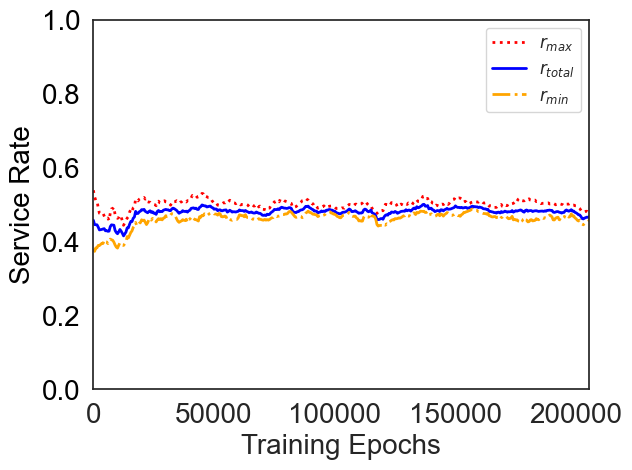}
		\caption{Learning with rates.}
	\end{minipage}
	\begin{minipage}[b]{0.45\linewidth}
		\centering
		\includegraphics[width=\linewidth]{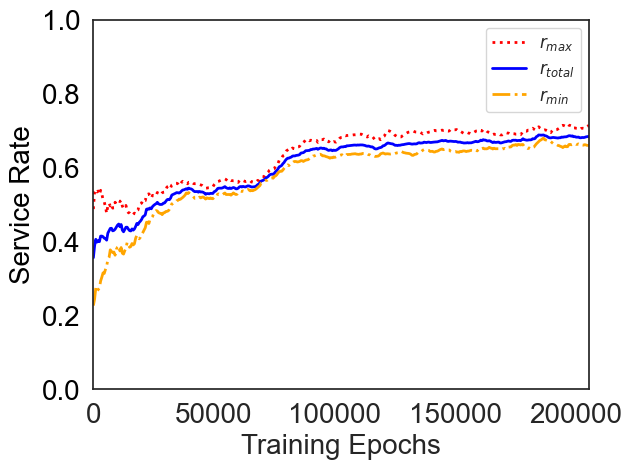}
		\caption{Learning with modified objective.}
	\end{minipage}
	\caption{Learning curves for learning with rates and learning with the modified objective for one of computational settings and $\alpha=0.5$.}
	\label{fig:modified_reward_learning_curves}
\end{figure}

For one of our computational settings, Figure~\ref{fig:modified_reward_learning_curves} presents a comparison for the learning curves with the original and modified reward function. 
On the left is the case for learning with rates, and on the right is for learning with the modified version. The horizontal axis is the number of training epochs, and the vertical axis is service rates. When optimizing with rates, it seems to learn only in the very early stage but starts to fluctuate with a flat trend overall. Within the same number of epochs, it significantly underperforms our proposed policy in both utility and fairness measures. Appendix~\ref{appendix:learning_rates} presents a detailed comparison for the two different reward functions implemented in different computational settings.  

In Appendix~\ref{appendix:prioritize}, we explore an alternative way of learning to ensure fairness by artificially increasing the reward for customers in an underrepresented area. Our results indicate that even though this method works relatively well, it requires domain knowledge as well as additional tuning and performs slightly worse than our proposed method.


\section{Computational Results}\label{sec:results}
In this section, we present a comprehensive computational study for the SDDFCS. In Section~\ref{sec:computational_settings}, we first introduce the computational settings. Section~\ref{benchmark} presents benchmarks as well as a baseline policy. In Section~\ref{sec:solution_quality}, we discuss solution quality including the trade-off between utility and fairness, fairness over time, and comparisons to the benchmark policies. In Section~\ref{sec:depot}, we analyze impact of different depot locations. In Section~\ref{sec:long_term}, we present long-term effect of ignoring fairness.

\subsection{Computational Settings}\label{sec:computational_settings}
Below we first describe the geographies where the setting of one region is inferior to the other, creating geographic unfairness. Note, in this paper, we consider two regions only, but our solution approach allows the number of regions $J$ to be any value. Then, we present the generation of customer requests, the vehicle fleet, and the setup of the neural net. 

\begin{figure}[h]
	\centering
	\includegraphics[width=0.5\textwidth]{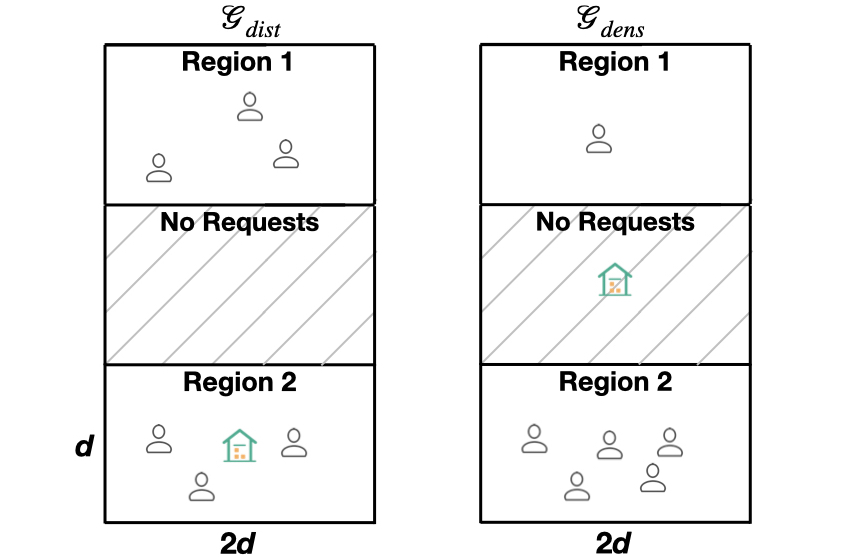}
	\caption{Illustration of the two geographies.}
	\label{fig:setting}
\end{figure}

\paragraph{Geographic settings.}
As mentioned earlier, low concentrations of memberships and long distance to the depot are the main difficulties that prevent companies such as Amazon from offering fair services \citep{ingold_2016}. These difficulties are also discovered in \citet{deepQ}. To address the industrial challenges, we test our solution approach using two geographic settings that are illustrated in Figure~\ref{fig:setting}. 

\begin{itemize}
    \item $\mathcal{G}$\textsubscript{dist}: In this setting, regions differ in relative distance to the depot. This setting is compromised of three equal-sized regions of width $d$ and length $2d$. The depot is located in the center of Region~$2$. The center of Region~$1$ is $2d$ away from the depot. The arrival rate of customer requests in both regions are $\lambda_1=\lambda_2=250$.
    \item $\mathcal{G}$\textsubscript{dens}: In this setting, Region~$1$ has lower population than Region~$2$ while both regions have the same relative distance to the depot. The arrival rates of customer requests are $\lambda_1=100$ and $\lambda_2=400$. The depot is located in the middle, between the two regions.
\end{itemize}

Both of them are designed to create circumstances that, without consideration of fairness, favor Region 2 over Region 1.

\paragraph{Requests.}
We assume the vehicle drivers work from $8$ am to $4$ pm, thus $t\textsubscript{max}=8$ hours. Customers make delivery requests from $8$ am to $3$ pm. Accepted requests must be serviced within $\bar{\delta}=4$ hours. All customer requests are generated according to a Poisson process. There are $500$ customer requests expected each day. The original $x$ and $y$ coordinates of customer locations are generated from independent and identical normal distributions. The standard deviation is $3$ km for each coordinate in the original data. With this value, about $50$\% of the customers reside in a core that is within $10$-minute vehicle travel time from the center, and about $99.9$\% of the customers are within a $30$-minute drive. These coordinates are then adjusted to reflect the \enquote{No Requests} area in the two geographies. 
\paragraph{Vehicles.}
To explore the effect of delivery resource availability, we test two different fleet sizes, three and five vehicles in the fleet. 
The vehicles travel at a speed of $30$km/h. To reflect the effect of road distances and traffic, we transform Euclidean distances to travel times using the method introduced in \cite{boscoe}. It takes $t_{\mathcal{N}}=3$ minutes to load packages at the depot and $t_{\mathcal{C}}=3$ minutes to drop off a package at a customer. 

The two unfair geographies are paired with three or five vehicles, resulting in four geography-fleet combinations, $\mathcal{G}\textsubscript{dist}-3v$, $\mathcal{G}\textsubscript{dist}-5v$, $\mathcal{G}\textsubscript{dens}-3v$, $\mathcal{G}\textsubscript{dens}-5v$. 

\paragraph{Training settings.}

The neural net consists of two hidden layers, each of which consists of $50$ neurons. We use ReLU as the activation function on the input and hidden layers. The output layer consists of $M+1$ nodes, where $M$ of them correspond to the value of accepting and assigning a customer to each vehicle, and the additional node represents the value of not offering service to this customer. We implement an $\epsilon$-greedy strategy for selecting actions during training, where $\epsilon$ exponentially decays from $1$ to $0.01$ during training. Experience replay is implemented to overcome the correlations in consecutive states \citep{lin}. 

We conduct the experiments on a training set of $1500$ instances that are sampled with replacement and a test set of $500$ instances. Here, an instance is a set of customer requests (request time and location) for one day. For each geography-fleet combination, we train the policies with different values of $\alpha$ from $\{0, 0.125, 0.25, 0.375, 0.5, 0.625, 0.75, 0.875, 1\}$. The value $\alpha=0$ represents the baseline policy where the fairness is not considered. Larger values of $\alpha$ place more emphasis on fairness during training. Training terminates after $200000$ epochs. After the training terminates, we evaluate the last ten policies on the testing set and calculate the expected utility and acceptance rates. 


\subsection{Benchmarks}\label{benchmark}
In this section, we introduce three policies to which we compare the performance of the proposed DQL policies with fairness consideration. The first one is called the \textit{baseline policy} (or policy $\alpha=0$). It is trained using the settings in Section~\ref{sec:computational_settings} with $\alpha=0$ and thus solely utility-driven. The other two policies incorporate fairness in decision making in different fashions, referred to as \textit{benchmark policies}. 
The two benchmarks are based on policy-function approximation (PFA). PFAs are policies that are generally characterized by a tuneable parameter. \citet{powell2011approximate} provides an overview of PFAs.

In the following, we describe how the two benchmark policies work. 
\begin{itemize}
    \item Bucket policy $\pi\textsuperscript{Bucket}$:
    
    The bucket policy seeks to achieve fairness by constraining the maximum service rate in any region and thus forcing service in less served region when a region reaches the limit. The $\pi\textsuperscript{Bucket}$ uses an accumulated-acceptance-rate threshold $\kappa_B$, $0\leq\kappa_B \leq 1$ to determine whether a request in a region should be accepted. We define the accumulated-acceptance rate of a region as the acceptance rate for that region from the beginning of the day to the current decision point. In $\pi\textsuperscript{Bucket}$, we use the same routing heuristic as that in Section~\ref{heuristic}. In the case there are multiple vehicles that can feasibly serve a request, the vehicle that has the smallest increase in travel time will be assigned. 

    The $\pi\textsuperscript{Bucket}$ works as follows. In the first $30$ minutes of the day, customers from both regions will be offered service if it is feasible, and no service otherwise. The consideration of this $30$-minute window is to ensure a large enough denominator when calculating the temporal acceptance rate. Thus, it avoids bad decisions such as offering no service to requests made in the early morning even if the vehicles are available. 
    
    After the first $30$ minutes, upon receiving a request $c_k$ from the region $j$, the bucket policy automatically offers no service if no vehicles can feasibly serve it. In the case it is feasible, the dispatcher compares the accumulated acceptance rate of the region $j$ to the threshold and then makes the acceptance decision, expressed as 
    \begin{equation}
    a_k=\left\{
    \begin{array}{ccl}
        1 &\ \ & \textrm{if } \frac{\sum_{0\leq l\leq k-1,c_l\in\mathcal{Z}_j} a_l}{\big|\{c_l:0\leq l\leq k-1,c_l\in\mathcal{Z}_j\}\big|}< \kappa_B,\\
             0    &\ \ & {\textrm{otherwise.}}
    \end{array} \right. 
    \end{equation}
    To find the best threshold $\kappa_B$ for each geography-fleet combination, we apply a search procedure that starts with a very small value of $\kappa_B$ and iteratively pushes it towards the upper bound $1$ until the minimum regional service rate cannot be improved anymore.
    
    \item Reserved-vehicle policy $\pi\textsuperscript{Reserved}$:
    
    Another PFA-based benchmark that we consider is the reserved-vehicle policy. The idea is to reserve a percentage of vehicles for the potential unfairly-treated region. This policy focuses on managing the fleet while the bucket policy focuses on making the acceptance decision. In the reserved-vehicle policy, a threshold $\kappa_R$ controls the number of vehicles that are exclusively reserved for Region~$1$. If there are $M$ vehicles in the fleet, then some number $\kappa_R\in\{1,2,\dots M-1\}$ of them serve customers only in Region~$1$ and the other $M-\kappa_R$ vehicles serve customers only in Region~$2$. We accept a request if it is feasible to do so, and offer no service otherwise. 
\end{itemize}

\subsection{Solution Quality}\label{sec:solution_quality}

In this section, we first compare the proposed DQL policies to the baseline policy. The analysis helps us gain insights into the trade-off between utility and fairness as well as fairness over time. Then, we discuss the two benchmarks.

\subsubsection{Comparison to Utility-Driven Baseline ($\alpha=0$).}\label{sec:tradeoff}

\begin{figure}[h!]
    \centering
	\begin{minipage}[b]{0.48\linewidth}
		\centering
		\includegraphics[width=\linewidth]{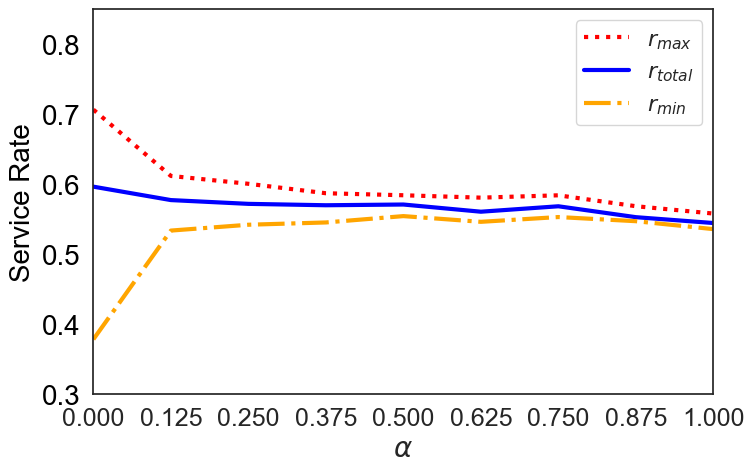}
		\caption{Service rates vs. values of $\alpha$.}
	\end{minipage}
	\begin{minipage}[b]{0.45\linewidth}
		\centering
		\includegraphics[width=\linewidth]{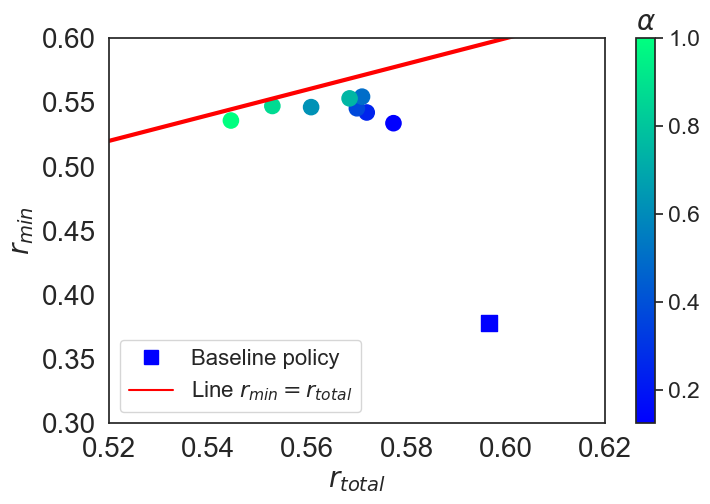}
		\caption{Pareto plot, $r$\textsubscript{total} vs. $r$\textsubscript{min}.}
	\end{minipage}
	\caption{Performance of the baseline and DQL policies over all geography-fleet combinations.}
	\label{trade_off}
\end{figure}

\paragraph{Utility-fairness trade-off.} We first quantify the trade-off between utility and fairness for the policies trained with different values of $\alpha$. 
For each geography-fleet combination, we evaluate the model on the test set and calculate the utility as well as the acceptance rates of all regions. For each value of $\alpha$, Figure~\ref{trade_off} presents the average results of the four combinations. 


In Figure~\ref{trade_off}(a), the horizontal axis plots the different values of $\alpha$, and the vertical axis represents the service rates $r\textsubscript{total}$ (utility), $r\textsubscript{min}$ (fairness), and $r\textsubscript{max}$. We refer the reader to Appendix~\ref{sec:performance} for the results of each geography-fleet combination. Not surprisingly, the baseline policy $\alpha=0$ serves the most requests but achieves both the smallest $r\textsubscript{min}$ but also the largest gap, about $34\%$, between $r\textsubscript{max}$ and $r\textsubscript{min}$. As $\alpha$ increases, the policies become fairer in the value of $r\textsubscript{min}$ at some cost in the utility. The results are in line with the intuition that there is a natural trade-off between utility and fairness. When $\alpha\leq0.5$, both $r\textsubscript{total}$ and $r\textsubscript{min}$ move monotonically, indicating the approach is efficient in lifting $r\textsubscript{min}$ up to alleviate unfairness. After $\alpha=0.5$, as more emphasis is placed on fairness, we observe that although the differences between $r\textsubscript{max}$ and $r\textsubscript{min}$ are still small, there are fluctuations in both $r\textsubscript{total}$ and $r\textsubscript{min}$. 

Figure~\ref{trade_off}(b) plots the corresponding Pareto set of the policies that are presented in Figure~\ref{trade_off}(a). Their color is determined by the value of $\alpha$ used for training. The horizontal axis is $r\textsubscript{total}$, and the vertical axis is $r\textsubscript{min}$. Note that, due to the dynamism and stochasticity of the SDDFCS, it is not realistic to directly find the Pareto set. Rather, in this section, we use the concept of Pareto as an evaluation metric. By definition, $r\textsubscript{min}$ can never exceed $r\textsubscript{total}$ in the problem. This boundary is represented by the red line. 

The Pareto plot shows that, when fairness is not considered, the utility-driven baseline policy $\alpha=0$ is far away from the boundary, indicating large room to improve the fairness performance. As the value of $\alpha$ increases from zero, the distance between the corresponding policy and the boundary is shortened. As the $\alpha$-value approaches one, corresponding polices are very close to the boundary, and thus there is not much space to further improve $r\textsubscript{min}$. On the other hand, as less and eventually none emphasis is put on utility, we see a declining $r\textsubscript{total}$. 
Our results show that, for the SDDFCS, a relatively balanced weight performs better than an imbalanced one. Thus, we focus on the policy $\alpha=0.5$ throughout the rest of the paper.

Notably, compared to such a highly utility-efficient policy, our policies that incorporate fairness in customer service suffer only a small loss in utility. For example, in $\mathcal{G}\textsubscript{dens}$ with three vehicles and $\alpha=0.5$, our policy improves $r\textsubscript{min}$ from $0.05$ to $0.46$ but only loses $6.2\%$ of the requests the baseline policy $\alpha=0$ serves.

\begin{figure}[h!]
    \centering
	\begin{minipage}[b]{0.48\linewidth}
		\centering
	\includegraphics[width=\textwidth]{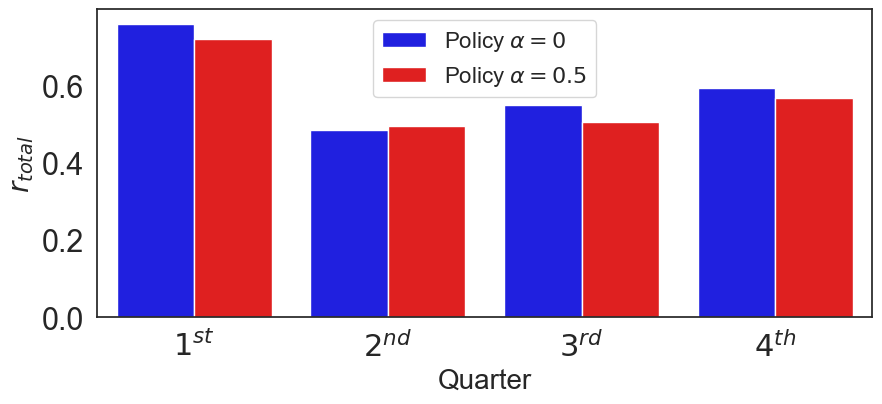}
	\caption{Utility during each quarter of the day.}
	\end{minipage}
	\begin{minipage}[b]{0.48\linewidth}
		\centering
	\includegraphics[width=\textwidth]{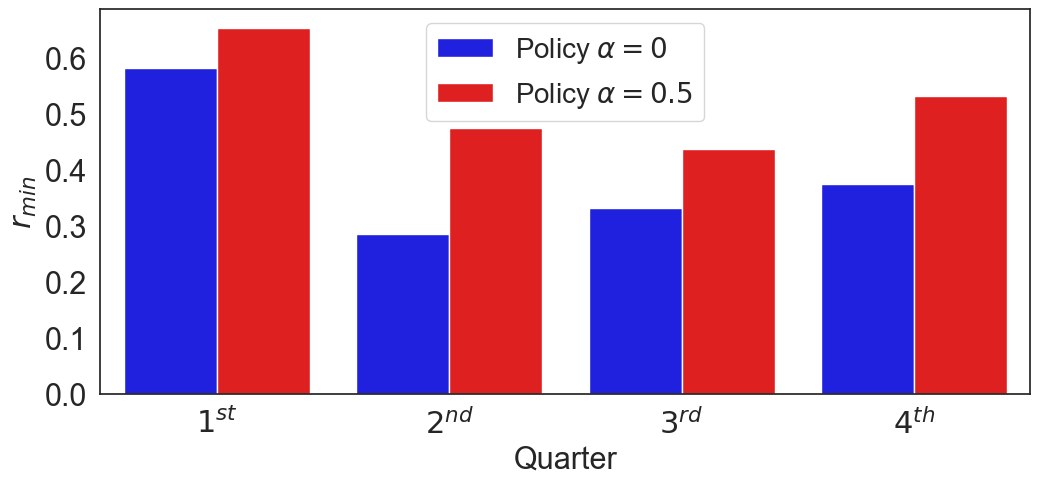}
	\caption{Fairness during each quarter of the day.}
	\end{minipage}
	\caption{Utility-fairness performance averaged over all geography-fleet combinations for policies $\alpha=0$ and $0.5$.}
	\label{fig:temporal_fairness}
\end{figure}

\paragraph{Fairness over time.} Our analysis so far demonstrates the effectiveness of our method in reducing spatial unfairness in SDD service. As our modified objective draws on incremental improvement in fairness over the course of a day, in this section, we will investigate if the policies learned by the proposed method achieve better temporal fairness. 
To do so, we divide the horizon of a day into four quarters and calculate the service rates during each quarter. Figure~\ref{fig:temporal_fairness} presents the average results for policies trained with $\alpha=0$ or $0.5$. 

Results show that our approach can alleviate the unfairness in SDD not only spatially but temporally throughout a day. Figure~\ref{fig:temporal_fairness}(a) presents the utility $r\textsubscript{total}$ throughout the day for both policies. Overall, the two policies show similar behaviors in the overall service rate over the day. They both accept more requests in the first quarter when the vehicles are relatively free. Later in the second and third quarters, fewer requests are accepted because vehicles are out serving the customers that are accepted in the first quarter. In the last quarter, when most of the assigned deliveries are completed, relatively more delivery resources are available, resulting in a slightly increased service rate.

Despite similar values in utility, the two policies show different performance with regard to fairness. Figure~\ref{fig:temporal_fairness}(b) presents $r\textsubscript{min}$ for each quarter. In the baseline policy $\alpha=0$, the unfairness gets enhanced starting the second quarter of the day. This implies that customers living in Region~$1$ (blue bars) have to place their order as early as possible to have a better chance to receive the service than those in Region~$2$. When fairness is considered, the value of $r\textsubscript{min}$ is significantly improved particularly in the last three quarters of the day when the delivery resources are even more constrained compared to the first quarter. Thus, while our model focused on spatial fairness, it also reduced the temporal unfairness over the course of the day.

\subsubsection{Comparison to Benchmarks with Consideration of Fairness.}

To compare the performance of the DQL policy and the benchmarks, for each policy and each geography-fleet combination, we find the best objective value calculated using Equation~\eqref{eq:fake_objective} 
with $\alpha=0.5$. The performance of a policy is then defined as the average over the four (combinations) objective values. 

\begin{figure}[h!]
    \centering
	\begin{minipage}[b]{0.55\linewidth}
		\centering
		\includegraphics[width=\linewidth]{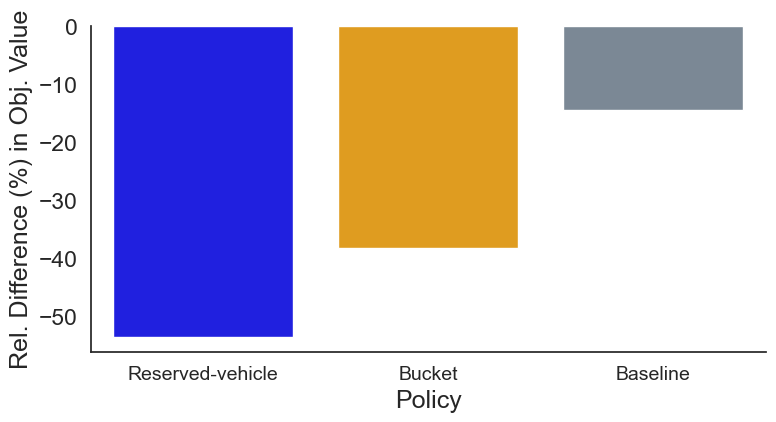}
	\end{minipage}
	\caption{Percent difference in average objective value relative to DQL policy $\alpha=0.5$.}
	\label{fig:benchmark_obj}
\end{figure}
For each benchmark as well as the baseline, Figure~\ref{fig:benchmark_obj} presents the percent difference relative to our proposed DQL policy. A negative value indicates an inferior performance of the benchmark or baseline policy. Overall, our policy outperforms the baseline and all the benchmark policies when coping with different geographies and fleet sizes. 
In addition to the average results in Figure~\ref{fig:benchmark_obj}, Table~\ref{table:benchmark} presents the detailed results for each benchmark, with bold numbers for the policies proposed in this paper.
\begin{table}[h]
    \scriptsize
	\centering
	\begin{tabular}{ccccccccc}
	\toprule
		\addlinespace[0.5em]
		\multicolumn{1}{c}{} &
		\multicolumn{4}{c}{$\mathcal{G}\textsubscript{dist}$}  &
		\multicolumn{4}{c}{$\mathcal{G}\textsubscript{dens}$}  \\
		\addlinespace[0.5em]
		\cmidrule{1-9}
		\addlinespace[0.5em]
         Fleet Size  & Parameter & Utility   & $r_1$   & $r_2$  &Parameter & Utility   & $r_1$   & $r_2$  \\
        \addlinespace[0.5em]
		\cmidrule{1-9}
		\addlinespace[0.5em]
        \multicolumn{9}{c}{Performance of $\pi\textsuperscript{Bucket}$ (parameter $\kappa_B$)}\\
		\cmidrule{1-9}
		$3$ vehicles  &  0.25 & 121.5 & 0.24  & 0.25 & 0.30 & 140.3 & 0.28 & 0.28\\ 
		$5$ vehicles   & 0.41  & 202.7 & 0.40 & 0.41 & 0.48 & 234.5 & 0.47 & 0.47\\ 
        \cmidrule(l){1-9}
        \addlinespace[0.5em]
        \multicolumn{9}{c}{Performance of $\pi\textsuperscript{Reserved}$ (parameter $\kappa_R$)}\\
        \cmidrule(l){1-9} 
3 vehicles      & 1   &130.4 &  0.08 & 0.44 & 1& 126.8 & 0.15 & 0.28  \\
           & 2   &110.1 &  0.28 & 0.15 &2 & 115.7 & 0.70 & 0.10  \\
5 vehicles      & 1   &255.5 &  0.10 & 0.92 &1 &258.8 & 0.16 & 0.61  \\
           & 2   &181.7 &  0.17 & 0.56 & 2&179.9 & 0.30 & 0.37  \\
           & 3   &164.8 &  0.35 & 0.30 & 3&170.0 & 0.79 & 0.22  \\
           & 4   &204.6 &  0.65 & 0.17 & 4&152.7 & 1.00 & 0.12   \\ 		
        \cmidrule(l){1-9} 
        \addlinespace[0.5em]
        \multicolumn{9}{c}{Performance of DQL policies (parameter $\alpha$)}\\
        \cmidrule(l){1-9} 
3 vehicles      & \multicolumn{1}{l}{0 (baseline)} & 231.8 & 0.27   & 0.66 & \multicolumn{1}{l}{0 (baseline)} &248.5  &0.05   & 0.62  \\
                & \multicolumn{1}{l}{0.5}   & \textbf{219.6} & \textbf{0.43}   & \textbf{0.45} &\multicolumn{1}{l}{0.5} &  \textbf{233.1}  &\textbf{0.46}   & \textbf{0.47}  \\

5 vehicles      & \multicolumn{1}{l}{0 (baseline)}   & 342.9   & 0.61 & 0.76 & \multicolumn{1}{l}{0 (baseline)}&369.8     &0.59   &  0.78 \\
                & \multicolumn{1}{l}{0.5}   &\textbf{341.1}    &\textbf{0.66}    &\textbf{0.71}  & \multicolumn{1}{l}{0.5}&\textbf{348.3}     &\textbf{0.66}   &  \textbf{0.71} \\
           \bottomrule
           \addlinespace[0.2em]
    \end{tabular}
	\caption{Detailed solution quality for all policies by geography-fleet combination.}
	\label{table:benchmark}
\end{table}

Policy $\pi\textsuperscript{Bucket}$ achieves small differences between the acceptance rates across regions because it strictly follows a rule designed to do so. However, its performance in both utility and fairness is inferior to our policies for every geography-fleet combination. For example, for $\mathcal{G}\textsubscript{dist}$ with three vehicles, our policy trained with $\alpha=0.5$ serves $81\%$ more requests than $\pi\textsuperscript{Bucket}$ when the $r\textsubscript{min}$ is $0.19$ higher than $\pi\textsuperscript{Bucket}$. It is worth noting that our DQL policies are not trained to shorten the gap between $r\textsubscript{max}$ and $r\textsubscript{min}$. However, when evaluated using such a metric, our policy still shows a comparable difference of $0.02$ to that of $0.01$ for $\pi\textsuperscript{Bucket}$. 

While easy to train and implement, $\pi\textsuperscript{Bucket}$ considers the data only from the past in the day and does not look ahead to the future when deciding whether to offer service. However, the approach proposed in this paper does so by anticipating the increases in both overall and minimal regional service rates. The results demonstrate the superiority of our proposed approach in learning a better utility as well as fairness compared to $\pi\textsuperscript{Bucket}$. 

Overall, $\pi\textsuperscript{Reserved}$ underperforms our proposed policy when both utility and fairness are considered. With three vehicles, the best performance of $r\textsubscript{min}$ over all the four settings is only $0.15$, about $0.30$ less than that under our policy, and the utility is far lower as well. With five vehicles, $\pi\textsuperscript{Reserved}$ is able to improve the best $r\textsubscript{min}$ to $0.30$, but there is still a difference of $-0.36$ compared to our policy. 
The results demonstrate the value of managing the fleet as a whole as we do in the proposed approach.

\subsection{Impact of Depot Location}\label{sec:depot}

Companies can selectively locate the depot to create a less unfair geographic setting, but there often exist restrictions in the real world that prevent them from doing so. On the other hand, if a depot already exists in a service area, it is usually not practical to knock it down and/or build another one. In this section, we will investigate whether our approach maintains its effectiveness with different depot locations and by how much it can ``save'' bad locations. Additionally, we can also gain insights from the analysis into where service providers should place a depot given the choice.

In previous sections, the results for $\mathcal{G}\textsubscript{dist}$ have shown that, when the bias results from the distance to the depot, our method can improve fairness when the depot is not ideally placed in the middle of the regions. In this section, we focus on the other geography, $\mathcal{G}\textsubscript{dens}$, where the bias lies in different arrival rates of requests. In addition to the depot location presented in Figure~\ref{fig:setting}, we conduct experiments for four more locations and train policies with $\alpha=0$ and $0.5$ for each location.
\begin{figure}[h!]
	\centering
	\includegraphics[width=0.72\textwidth]{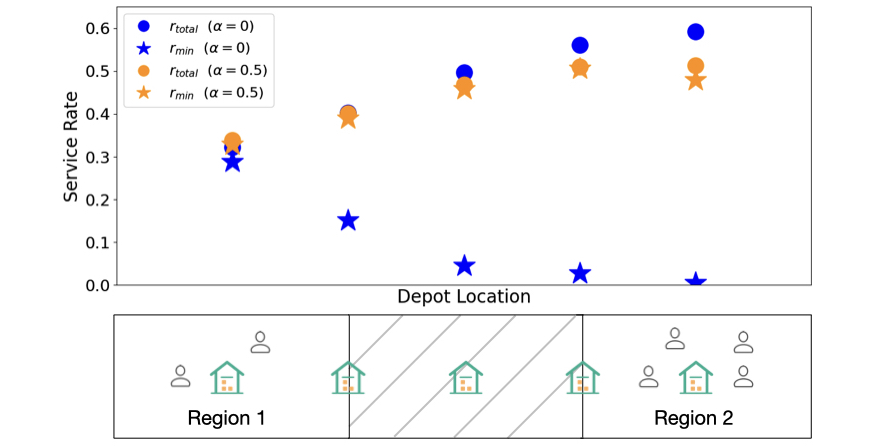}
	\caption{Utility-fairness performance for different depot locations in $\mathcal{G}\textsubscript{dens}$ with 3 vehicles.}
	\label{fig:depot_location}
\end{figure}

Figure~\ref{fig:depot_location} presents the results along with an illustration of the five depot locations. We plot the utility-fairness data in the upper sub-figure, and in the bottom of the figure is the geographic setting $\mathcal{G}\textsubscript{dens}$. To provide a better visual reference to the results, we rotate the map of $\mathcal{G}\textsubscript{dens}$ counterclockwise by $90$ degrees such that each depot in the upper figure corresponds to a location in the bottom figure. In the utility-fairness figure (the upper figure), we use dots (stars) to represent $r\textsubscript{total}$ ($r\textsubscript{min}$) with blue (orange) for policies trained with $\alpha=0$ ($\alpha=0.5$). 


\paragraph{Effectiveness with different depot locations.} When fairness is considered, results show that location of the depot matters in terms of utility-fairness trade-off. For utility of the policies $\alpha=0.5$, we observe a non-monotonic behavior---when the depot is moved from the left to right, the utility increases until the fourth position and then slightly decreases at the right-most one. This is because setting the depot even farther from Region~$1$ increases the time needed to serve customers in Region~$1$ and thus reduces the overall utility.

Overall, our proposed approach can achieve better $r\textsubscript{min}$ than the baseline at each depot location considered. Along with the results for $\mathcal{G}\textsubscript{dist}$, it exhibits the effectiveness of our method is valid with different depot locations, and thus it offers service providers the opportunity to achieve better fairness from any location. 



\paragraph{Impact on locating a new site.} Without our approach, to incorporate fairness, service providers need to locate the depot in the less-populated region, represented by the left-most blue dot and star in Figure~\ref{fig:depot_location}. However, doing so greatly sacrifices utility compared to our policies, the orange dots and stars on the right. This sacrifice results from the fact that it now becomes more costly in terms of time to serve customers in the region of higher density and thus fewer customers are served overall. Our results show that, when regions have different arrival rates, service providers should consider locating the depot closer to the region with more requests to achieve both high utility and fairness. Although it is beyond the scope of this work, this ideal location may potentially benefit businesses that integrate the depot and a brick-and-mortar store. Such a store would be relatively close to the densely populated region, making it easier for those customers to shop onsite while also providing a utility-fairness balance for delivery.

\subsection{Long-Term Effect of Ignoring Fairness}\label{sec:long_term}

As an emerging topic, fairness in commercial delivery has not been extensively studied yet. However, fairness-related work is available in other problem domains. \citet{hassan2013measuring} study the effect of service fairness in telecommunication sector. In their work, the authors show a direct effect of service fairness on customer loyalty and the need for companies to consider fairness to stay competitive. Thus, in this section, we investigate how ignoring fairness impacts the business in the long run. 

We take as an example the geography $\mathcal{G}\textsubscript{dist}$ with five vehicles in this section. We use two already-trained policies with $\alpha=0$ and $0.5$, respectively. For the trained policies, the baseline $\alpha=0$ accepts about $61\%$ of the requests from Region~$1$ and $76\%$ from Region~$2$. The policy $\alpha=0.5$ serves about $66\%$ from Region~$1$ and $71\%$ from Region~$2$. 

Yet, over the long term, customers will change their behavior according to their likelihood of being served. Notably, those who find themselves underserved will either lose interest in the service or move their business to competitors. To reflect the long term, we test the two policies on a period of one year. To mimic customers' reactions to different levels of service rate, at the end of every month, we calculate the regional service rates for the two regions and then update the arrival rate for the following month. 
Let $r\textsubscript{threshold}$ be a service-rate threshold. This threshold is equivalent to the \textit{reference point} or \textit{customer expectation} in the field of fairness research \citep{mathies2011role}, to which customers compare their perceived information. A value below the threshold will likely lead to customer churn while a value above might lead to increasing demand. To model this, the arrival rate for each region is updated as $\lambda\textsubscript{new}=\lambda\textsubscript{old} + \lambda\textsubscript{old}\cdot (r-r\textsubscript{threshold})$. With this update, fewer customers will request service if there has been a low service rate for the region lately, and vice versa. The increase or decrease in the arrival rate is proportional to how much the actual service rate deviates from $r\textsubscript{threshold}$. For $r\textsubscript{threshold}$, we consider different levels of customers' expectation on service rate, i.e., $r\textsubscript{threshold}\in\{60\%,65\%,70\%,75\%,80\%$\}. A lower threshold indicates that customers have a lower expectation of service.

\begin{figure}[h!]
    \centering
	\begin{minipage}[b]{0.4\linewidth}
		\centering
		\includegraphics[width=\linewidth]{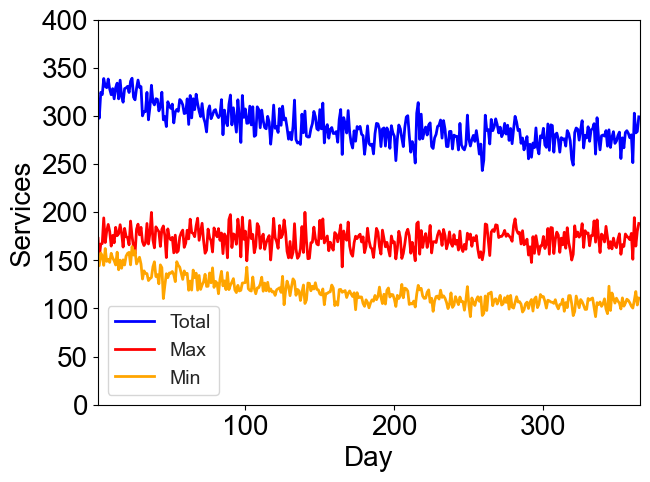}
		\caption{Baseline policy $\alpha=0$.}
	\end{minipage}
	\begin{minipage}[b]{0.4\linewidth}
		\centering
		\includegraphics[width=\linewidth]{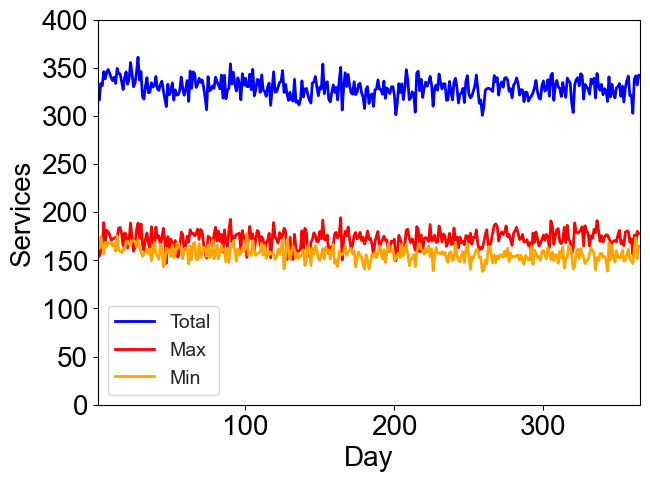}
		\caption{Policy $\alpha=0.5$.}
	\end{minipage}
	\caption{Average number of accepted requests per day over a year-long horizon for policies $\alpha=0$ and $0.5$ ($r\textsubscript{threshold}=70\%$).}
	\label{fig:long_term_70}
\end{figure}

Figure~\ref{fig:long_term_70} shows the actual number of accepted requests for both policies over the $1$-year period when $r\textsubscript{threshold}$ is $70\%$. The results for the other thresholds are presented in Appendix~\ref{append:ignore}. As shown in the figure, the policy $\alpha=0.5$ remains roughly unchanged in its performance over the year-long horizon, while the baseline loses customers in Region~$1$ and suffers a decrease in overall performance. Thus, even though the baseline initially received higher utility on the cost of fairness, after a year, both fairness and utility are below the values for the policy $\alpha=0.5$.

\begin{figure}[h!]
    \centering
	\begin{minipage}[b]{0.65\linewidth}
		\centering
		\includegraphics[width=\linewidth]{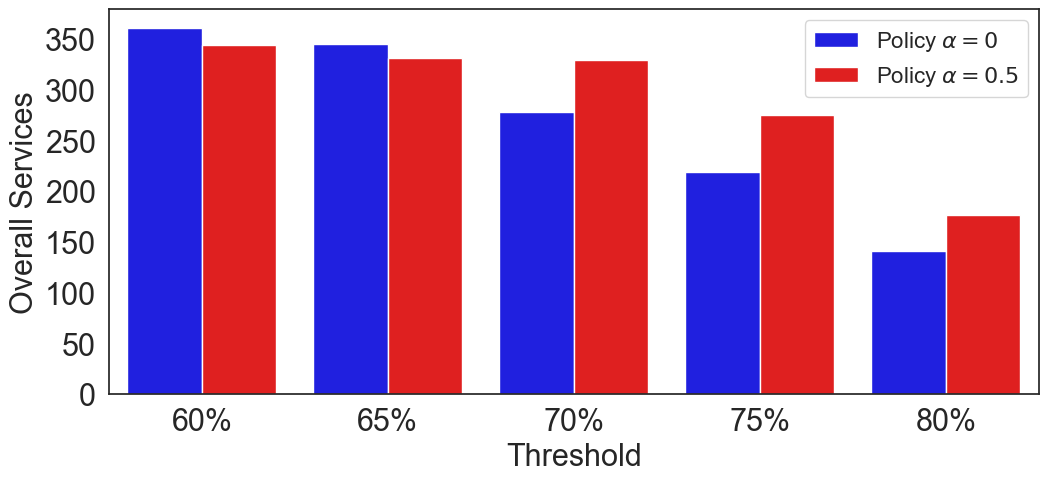}
	\end{minipage}
	\caption{Average number of accepted requests per day in the last month for policies $\alpha=0$ and $0.5$.}
	\label{fig:long_term}
\end{figure}

For the two policies and different service-level expectations, Figure~\ref{fig:long_term} presents the average number of accepted requests per day in
the last month of the horizon.
As expected, as the expectation on service level increases, both policies show a declining overall number of services. The baseline policy $\alpha=0$ starts to show a significant drop in the number of services when the threshold is $70\%$. However, we do not see any significant decrease for the policy $\alpha=0.5$ until the expectation reaches $75\%$. This shows that our policy is more capable of coping with high expectations on service level. Further, starting from the expectation of $70\%$, the policy $\alpha=0.5$ outperforms the baseline with regard to overall utility.

In summary, the analysis suggests that ignoring fairness in customer service will alienate customers in underserved regions and thus result in a long-term loss of overall revenue. When customers have high expectations on service level, our approach outperforms the baseline in profitability. The results echo with the findings in service literature---service
fairness has the highest overall effect on customer loyalty \citep{giovanis2015role} and satisfaction \citep{berry1995great}, which have been proved to be reliable sources for profitable growth \citep{hbs}.


\section{Conclusions and Future Work}\label{sec:conclusions}
In this paper, we present a SDD problem with fairness. In addition to the overall service rate, we maximize the minimum regional service rate across all regions. We develop a deep Q-learning solution approach for the problem. Due to the instability inherent in learning directly with the rates, we introduce a novel transformation of the objective that draws on incremental improvement throughout a day. 

Computational results demonstrate the effectiveness of the method in reducing the unfairness in SDD in various customer geographies. Remarkably, results show that our proposed method makes relatively small sacrifice in short-term utility. For example, the utility decreases by only $6.2\%$ when the minimum regional service rate is improved from $0.05$ to $0.46$ in one of our settings. This result suggests that companies do not need to pay a big price to achieve significant improvements in fairness. Our additional experiments also show that, when customers' expectation on service opportunities is considered, our method may not only achieve fairer outcomes but also better long-term financial outcomes. 
Further, our method provides valuable insights for companies in choosing depot locations. Without using a method like ours, in order to improve fairness, companies need to place the depot close to underserved region, which results in a significant cost of utility. With our model, the suggested depot location is close to or in the advantaged region, and fairness can be improved at a lower cost of utility. For existing depot locations that are far way from certain regions, such as those of Amazon's warehouses \citep{ingold_2016}, our approach can remedy the disadvantaged locations and offer fair services to customers. 

There are several directions for future work. First, there are situations, such as the COVID-19 pandemic, when certain groups of people that are of higher risk must limit social contact. Thus, there is value in studying how to prioritize some customers over others and how it affects the the decision making. Second, in addition to customer service, it is worth investigating whether our method works effectively in balancing the workload for drivers/couriers. It is also an interesting topic to achieve fairness in both sides at the same time. Finally, we can consider extensions of the SDDFCS, such as multiple depots serving multiple regions and/or a heterogeneous fleet making deliveries. 


\bibliography{arXiv_R3.bib}  
\bibliographystyle{abbrvnat}






\newpage
\begin{appendix}\label{append}

\section{Learning Directly with Rates}\label{appendix:learning_rates}

Table~\ref{table:learn_rates} presents a comparison for the two different reward functions performed on the four geography-fleet combinations. For both $r\textsubscript{total}$ and $r\textsubscript{min}$, we present the absolute difference relative to the policies trained with the modified objective. 

\begin{table}[h]
	\centering
	\begin{tabular}{ccccc}
	\toprule
		\multicolumn{1}{c}{} &
		\multicolumn{2}{c}{$\mathcal{G}\textsubscript{dist}$}  &
		\multicolumn{2}{c}{$\mathcal{G}\textsubscript{dens}$}  \\
		\addlinespace[0.2em]
		\cmidrule{1-5}
		\addlinespace[0.2em]
		\multicolumn{1}{c}{Fleet Size} &
		\multicolumn{1}{c}{Difference in $r\textsubscript{total}$}  &
		\multicolumn{1}{c}{Difference in $r\textsubscript{min}$}  &
		\multicolumn{1}{c}{Difference in $r\textsubscript{total}$}  &
		\multicolumn{1}{c}{Difference in $r\textsubscript{min}$}  \\
		\addlinespace[0.2em]
		\cmidrule{1-5}
		\addlinespace[0.05em]
		$3$ vehicles   & -0.11  & -0.12   & -0.10 & -0.11 \\ 
		\addlinespace[0.2em]
		$5$ vehicles   &  -0.20  & -0.20 & -0.06 & -0.03 \\ 
		\addlinespace[0.2em]
		\bottomrule
		\addlinespace[0.2em]
	\end{tabular}
	\caption{Performance of learning with rates relative to learning with modified objective.}
	\label{table:learn_rates}
\end{table}

In all the four geography-fleet combinations, we observe significant gaps in $r\textsubscript{total}$ compared to policies trained with the modified objective, ranging from $-0.20$ to $-0.06$. Similar gaps appear in $r\textsubscript{min}$ as well. For example, in $\mathcal{G}\textsubscript{dist}$ with three vehicles, the minimal regional service rate is $0.43$ when learning with the modified objective but only $0.32$ for learning directly with the rates, resulting in a difference of $-0.11$. The comparison demonstrates that, although intuitive and straightforward, learning directly with the rates shows not only an undesired trend in learning curves but also lower values in both overall and minimal regional service rates.

\section{Prioritized-Customer Policy}\label{appendix:prioritize}

Computational results in \citet{deepQ} have demonstrated that DQL is capable of learning with a heterogeneous reward. Motivated by their work, we consider another benchmark policy $\pi\textsuperscript{Priority}$ called the prioritized-customer policy. 
    
The $\pi\textsuperscript{Priority}$ is similar to our DQL policies as both use a NN to approximate the value of state-action pairs. Rather than multiple objectives, $\pi\textsuperscript{Priority}$ learns to optimize the cost (reward) function of a single objective but values some customers more than others. Fairness is implicitly incorporated in the objective through these differentiated values. In addition to approximating Q-values, the parameter that controls the different priorities of customers makes $\pi\textsuperscript{Priority}$ also parametric CFA-based. 

The corresponding single-objective MDP model differs from the proposed multi-objective one only in the reward. Instead of a linear combination of utility and fairness rewards, the reward function in $\pi\textsuperscript{Priority}$ is concerned with prioritizing certain customers by assigning them a larger reward. Let $p_j$ be a notion of priority assigned for customers in the region $\mathcal{Z}_j$. Then, the reward of an action $x_k$ in the state $S_k$ is defined as
    \begin{equation}
    R(S_k,x_k)=\left\{
    \begin{array}{ccl}
    0      &\ & {\textrm{if } a_k=0,}\\
    p_j  &\ & {\textrm{if } a_k=1,}
    \end{array} \right. 
    \end{equation}
    provided the request $c_k$ is made from the region $\mathcal{Z}_j$. Note, if $p_j=1$ for all regions, then we have the benchmark policy ($\alpha=0$) introduced in Section~\ref{sec:tradeoff}. We then train $\pi\textsuperscript{Priority}$ with $J=2$ using the same settings described in Section~\ref{sec:computational_settings}. The reward for customers in Region~$2$ is always $p_2=1$. To increase the chance that customers in Region~$1$ receive the service, we technically set a larger reward for them chosen from $p_1\in\{1.25,1.5,1.75,2\}$. Clearly, this benchmark relies on a priori knowledge of what region is going to be the underserved region.

For $\pi\textsuperscript{Priority}$, we notice that it outperforms our proposed policy in $\mathcal{G}\textsubscript{dens}$ with certain values of $p_1$. For example, when it is trained with five vehicles and $p_1=1.5$, the policy serves more requests overall and a better minimal regional rate. It is possibly due to the small arrival rate of requests for Region~$1$. In $\mathcal{G}\textsubscript{dens}$, the expected arrival rate for Region~$1$ is $n_1=100$ requests per day. However, the actual number of requests varies from day to day. The effect of the variation is increased by the small value of $n_1$. For example, if the actual number of requests is $20$ less than the expected $100$ in a day, accepting a request from this region improves the regional service rate by $\frac{1}{80}=0.0125$, and the difference from using the expected value is $0.0125-\frac{1}{100}=0.0025$. The difference becomes only $\frac{1}{230}-\frac{1}{250}=0.0003$ if $n_1=250$. The increased effect of the variation could lead to the underperformance in this particular geography-fleet combination.

However, the approach proposed in this paper still shows an overall superiority over the $\pi\textsuperscript{Priority}$. First, in the other geography $\mathcal{G}\textsubscript{dist}$, the policies learned by our method significantly outperforms the $\pi\textsuperscript{Priority}$. It indicates that the proposed approach is more capable of coping with different types of bias existing in the system the superiority in the other geography. Second, the $\pi\textsuperscript{Priority}$ relies on the knowledge of what region is underserved and more importantly, requires the tuning of parameter $p_j$. In the case of multiple regions, the number of potential combinations of $p_j$'s increases exponentially. Nevertheless, our approach requires less to none tuning of parameter of $\alpha$.

\section{Performance of Policies}\label{sec:performance}

\begin{figure}[h]
    \centering
    \begin{minipage}[b]{0.4\linewidth}
        \centering
        \includegraphics[width=1\linewidth]{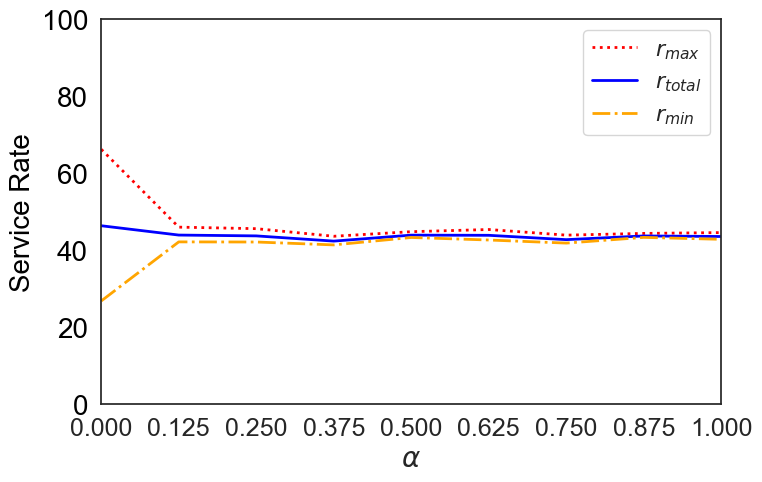} 
        \caption{$\mathcal{G}\textsubscript{dist}$, 3 vehicles}
    \end{minipage}
    \begin{minipage}[b]{0.4\linewidth}
        \centering
        \includegraphics[width=0.98\linewidth]{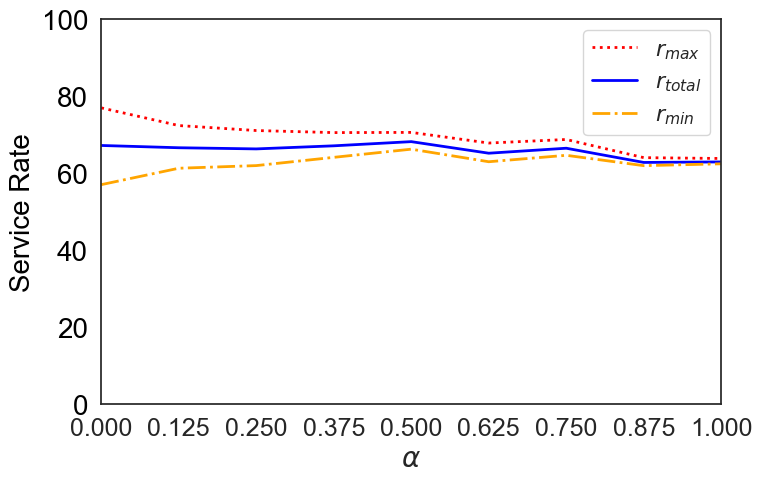} 
        \caption{$\mathcal{G}\textsubscript{dist}$, 5 vehicles}
    \end{minipage}
    \begin{minipage}[b]{0.4\linewidth}
        \centering
  	    \includegraphics[width=0.98\linewidth]{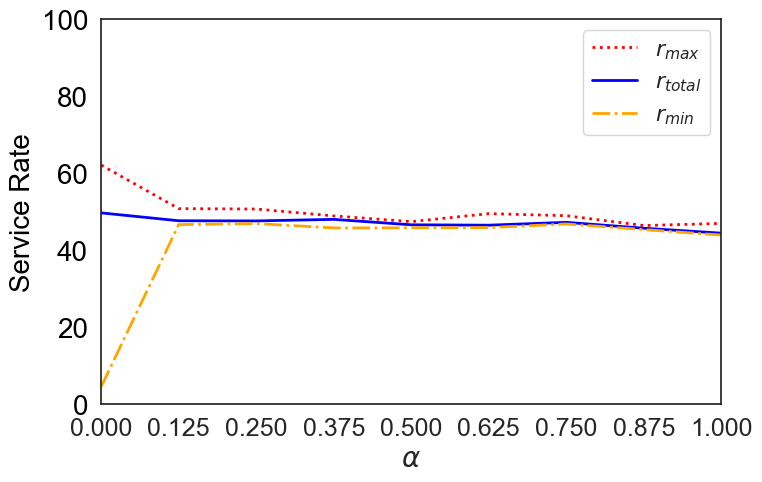} 
  	    \caption{$\mathcal{G}\textsubscript{dens}$, 3 vehicles}
    \end{minipage}
    \begin{minipage}[b]{0.4\linewidth}
        \centering
  	    \includegraphics[width=0.98\linewidth]{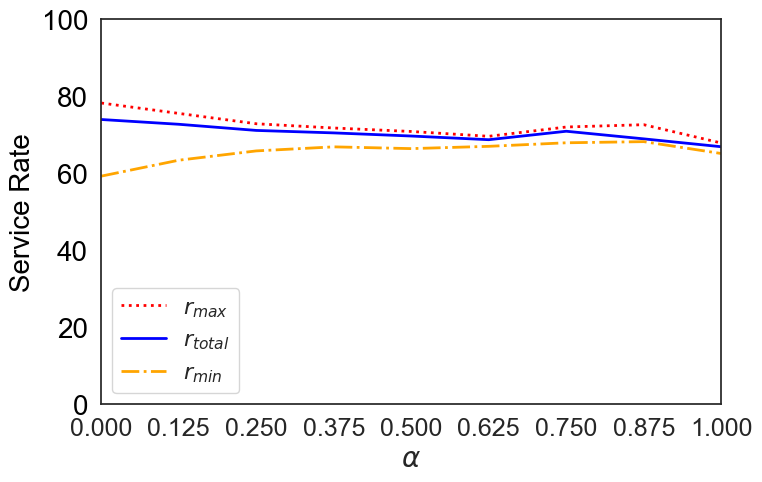} 
  	    \caption{$\mathcal{G}\textsubscript{dens}$, 5 vehicles}
    \end{minipage}
    \caption{Utility-fairness trade-off curves for different geography-fleet combinations.}
    \label{fig:performance}
\end{figure}

Figure~\ref{fig:performance} presents the performance for each geography-fleet combination discussed in Section~\ref{sec:computational_settings}.

Consistent with the average performance, as the value of $\alpha$ increases, the utility decreases as the minimal regional service rate increases.

\section{Impact of Ignoring Fairness}\label{append:ignore}
\begin{figure}[h!]
    \centering
	\begin{minipage}[b]{0.4\linewidth}
		\centering
		\includegraphics[width=\linewidth]{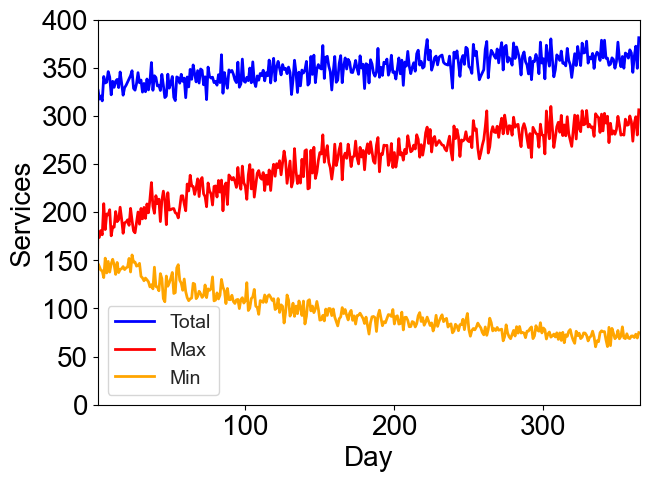}
		\caption{Benchmark policy.}
	\end{minipage}
	\begin{minipage}[b]{0.4\linewidth}
		\centering
		\includegraphics[width=\linewidth]{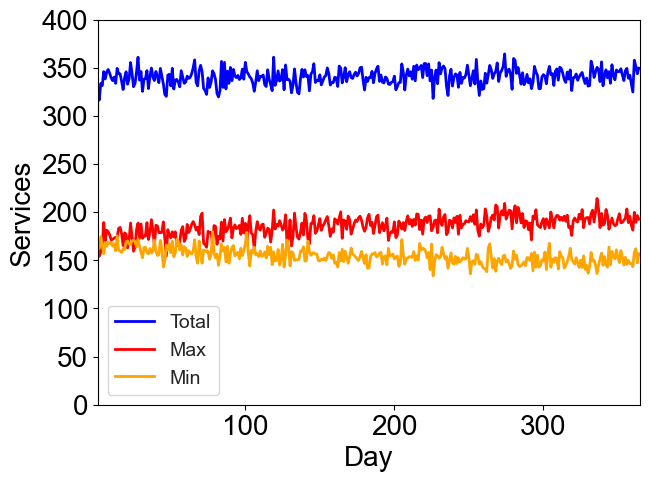}
		\caption{Fair policy.}
	\end{minipage}
	\caption{Number of accepted requests for benchmark and fair policies over a year for $r\textsubscript{threshold}=60\%$.}
	\label{fig:long_term_60}
\end{figure}

\begin{figure}[h!]
    \centering
	\begin{minipage}[b]{0.4\linewidth}
		\centering
		\includegraphics[width=\linewidth]{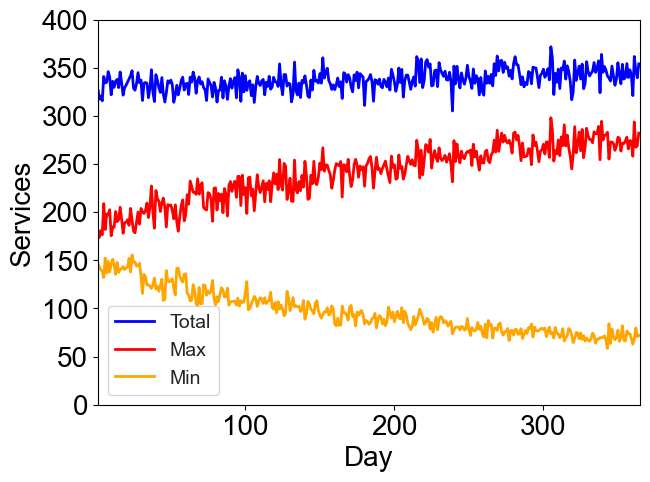}
		\caption{Benchmark policy.}
	\end{minipage}
	\begin{minipage}[b]{0.4\linewidth}
		\centering
		\includegraphics[width=\linewidth]{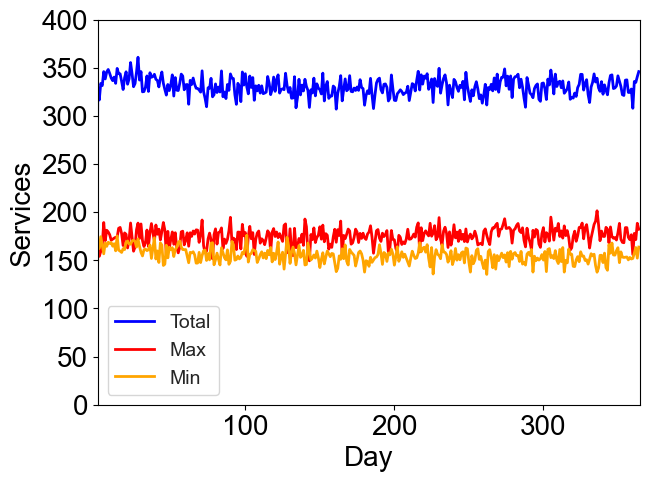}
		\caption{Fair policy.}
	\end{minipage}
	\caption{Number of accepted requests for benchmark and fair policies over a year for $r\textsubscript{threshold}=65\%$.}
	\label{fig:long_term_65}
\end{figure}

\begin{figure}[h!]
    \centering
	\begin{minipage}[b]{0.4\linewidth}
		\centering
		\includegraphics[width=\linewidth]{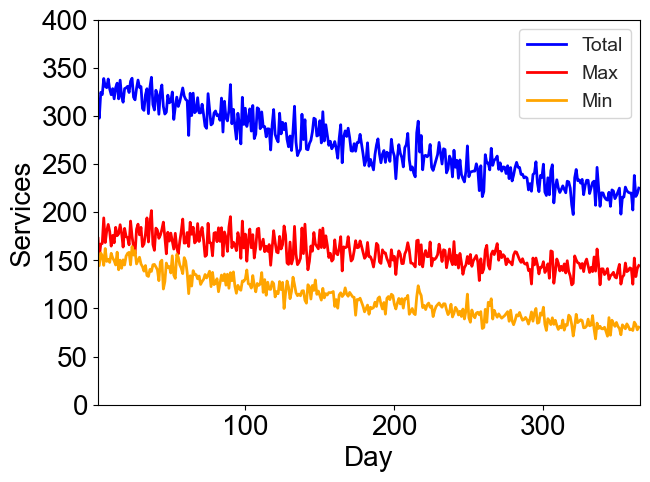}
		\caption{Benchmark policy.}
	\end{minipage}
	\begin{minipage}[b]{0.4\linewidth}
		\centering
		\includegraphics[width=\linewidth]{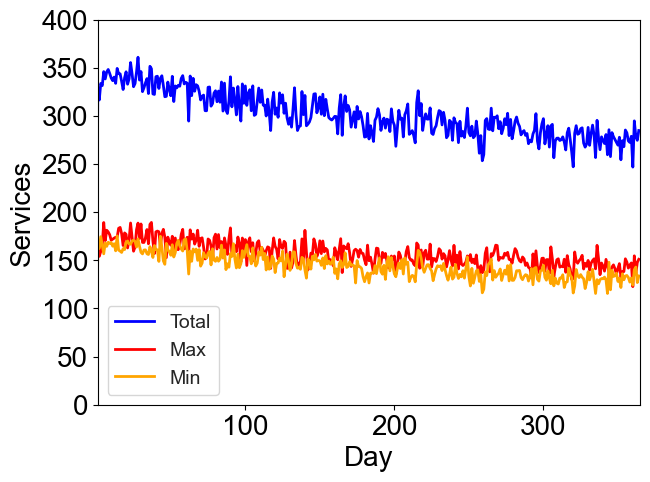}
		\caption{Fair policy.}
	\end{minipage}
	\caption{Number of accepted requests for benchmark and fair policies over a year for $r\textsubscript{threshold}=75\%$.}
	\label{fig:long_term_75}
\end{figure}

\begin{figure}[h!]
    \centering
	\begin{minipage}[b]{0.4\linewidth}
		\centering
		\includegraphics[width=\linewidth]{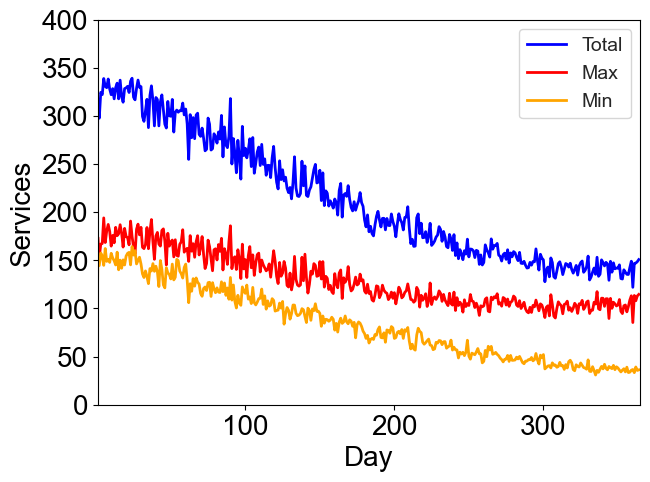}
		\caption{Benchmark policy.}
	\end{minipage}
	\begin{minipage}[b]{0.4\linewidth}
		\centering
		\includegraphics[width=\linewidth]{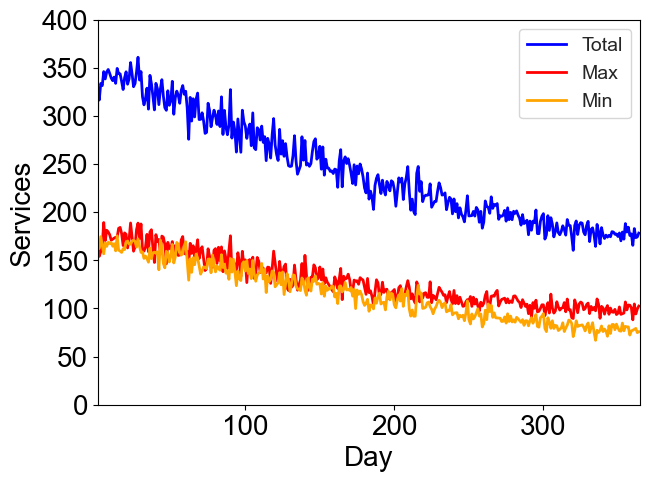}
		\caption{Fair policy.}
	\end{minipage}
	\caption{Number of accepted requests for benchmark and fair policies over a year for $r\textsubscript{threshold}=80\%$.}
	\label{fig:long_term_80}
\end{figure}

Figures~\ref{fig:long_term_60} to~\ref{fig:long_term_80} present the number of accepted requests for policies $\alpha=0$ and $\alpha=0.5$ over the $1$-year period with the other values of $r\textsubscript{threshold}$.

When $r\textsubscript{threshold}=60\%$ or $70\%$, the baseline policy $\alpha=0$ serves more customers overall. However, doing so is at the expense of losing customers from the already-underserved region. The performance of the policy $\alpha=0.5$ is roughly unchanged. When $r\textsubscript{threshold}=75\%$ or $80\%$, both policies see a decrease in the total number of requests served. Notably, under the policy $\alpha=0.5$, the difference in the number of services between the two regions remains small, even when the overall number of services declines. 

\end{appendix}

\end{document}